\documentclass[dvipsnames]{article}

\usepackage{arxiv}

\usepackage[utf8]{inputenc} 
\usepackage[T1]{fontenc}    
\usepackage{hyperref}       
\usepackage{url}            
\usepackage{booktabs}       
\usepackage{amsfonts}       
\usepackage{nicefrac}       
\usepackage{microtype}      
\usepackage{graphicx}
\usepackage{doi}
\usepackage{todonotes}  
\usepackage{amsmath, amssymb}
\usepackage{paralist} 
\usepackage{colortbl,xcolor} 
\usepackage{tikz} 
\usepackage{makecell} 
\usepackage[numbers]{natbib} 

\newcommand*\circled[1]{\tikz[baseline=(char.base)]{    \node[shape=circle,draw,fill=SpringGreen,inner sep=2pt,font=\sffamily\scriptsize] (char) {#1};}}

\usepackage{enumitem} 
\usepackage{tabularx} 
\usepackage{multirow} 
\title{Balancing Autonomy and Alignment:\\ A Multi-Dimensional Taxonomy for Autonomous LLM-powered Multi-Agent Architectures}

\author{\href{https://orcid.org/0000-0002-0589-204X}{\includegraphics[scale=0.06]{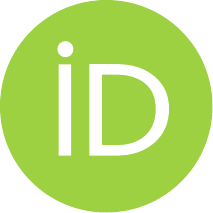}\hspace{1mm} Thorsten Händler}\\
    Ferdinand Porsche Mobile University of Applied Sciences (FERNFH)\\
	Wiener Neustadt, Austria \\
	\texttt{thorsten.haendler@fernfh.ac.at} \\
}

\date{}

\renewcommand{\headeright}{}
\renewcommand{\headerleft}{\sffamily \scriptsize Balancing Autonomy and Alignment}
\renewcommand{\undertitle}{}
\renewcommand{\shorttitle}{}

\hypersetup{
pdftitle={Balancing Autonomy and Alignment: A Multi-Dimensional Taxonomy for Autonomous LLM-powered Multi-Agent Architectures},
pdfsubject={cs.AI, cs.SE, cs.MA},
pdfauthor={Thorsten Händler},
pdfkeywords={Taxonomy, autonomous agents, multi-agent collaboration, large language models (LLMs), AI systems classification, alignment, software architecture, architectural viewpoints, software-design rationale, context interaction, artificial intelligence, domain-ontology diagram, feature diagram, radar chart.},
}

\begin{document}
\maketitle

\begin{abstract}
Large language models (LLMs) have revolutionized the field of artificial intelligence, endowing it with sophisticated language understanding and generation capabilities. However, when faced with more complex and interconnected tasks that demand a profound and iterative thought process, LLMs reveal their inherent limitations. Autonomous LLM-powered multi-agent systems represent a strategic response to these challenges. Such systems strive for autonomously tackling user-prompted goals by decomposing them into manageable tasks and orchestrating their execution and result synthesis through a collective of specialized intelligent agents. Equipped with LLM-powered reasoning capabilities, these agents harness the cognitive synergy of collaborating with their peers, enhanced by leveraging contextual resources such as tools and datasets.
While these architectures hold promising potential in amplifying AI capabilities, striking the right balance between different levels of autonomy and alignment remains the crucial challenge for their effective operation. 
This paper proposes a comprehensive multi-dimensional taxonomy, engineered to analyze how autonomous LLM-powered multi-agent systems balance the dynamic interplay between autonomy and alignment across various aspects inherent to architectural viewpoints such as goal-driven task management, agent composition, multi-agent collaboration, and context interaction. 
It also includes a domain-ontology model specifying fundamental architectural concepts. Our taxonomy aims to empower researchers, engineers, and AI practitioners to systematically analyze, compare, and understand the architectural dynamics and balancing strategies employed by these increasingly prevalent AI systems, thus contributing to ongoing efforts to develop more reliable and efficient solutions.
The exploratory taxonomic classification of selected representative LLM-powered multi-agent systems illustrates its practical utility and reveals potential for future research and development.
\end{abstract}


\section{Introduction}
\label{sec:introduction}

In recent years, the emergence and the technological feasibility of large language models (LLMs) have revolutionized the field of artificial intelligence \cite{brown2020language,ouyang2022training,thoppilan2022lamda,chowdhery2022palm,zhang2022opt}. 
Pre-trained on vast amounts of text data, these models have catalyzed significant advancements by enabling sophisticated language understanding and generation capabilities, opening doors to a broad range of applications \cite{bommasani2021opportunities,bubeck2023sparks,kaddour2023challenges}. 
Yet, despite their remarkable capabilities, LLMs also have inherent limitations. 

While LLMs excel at generating outputs based on patterns identified in their training data, they lack a genuine understanding of the real world. Consequently, their outputs might seem plausible on the surface, but can be factually incorrect or even \textit{hallucinated} \cite{maynez2020faithfulness,ji2023survey}. 
Moreover, despite their proficiency in handling vast amounts of textual information and their rapid processing and pattern recognition capabilities, LLMs struggle with maintaining consistent logic across extended chains of reasoning. This deficiency hinders their ability to engage in a deliberate, in-depth, and iterative thought process (aka \textit{slow thinking}) \cite{sloman1996empirical,kahneman2011thinking,fabiano2023fast,lin2023swiftsage}. As a result, LLMs encounter difficulties when it comes to handling more complex and interconnected tasks \cite{kojima2022large,wei2022chain}.

These limitations of individual LLMs have led to the exploration of more sophisticated and flexible AI architectures including multi-agent systems that aim at accomplishing complex tasks, goals, or problems with the \textit{cognitive synergy} of multiple autonomous LLM-powered agents \cite{autogpt2023,babyagi2023,superagi2023,park2023generative,shen2023hugginggpt,li2023camel,agentgpt2023,hong2023metagpt}.
Such systems tackle user-prompted goals by employing a \textit{divide \& conquer} strategy, by breaking them down into smaller manageable tasks. These tasks are then assigned to specialized agents, each equipped with a dedicated role and the reasoning capabilities of an LLM, as well as further competencies by utilizing contextual resources like data sets, tools, or further foundation models.
Taking a cue from Minsky's \textit{society of mind} theory \cite{minsky1988society}, the key to the systems' problem-solving capability lies in orchestrating the iterative collaboration and mutual feedback between these more or less \textit{'mindless'} agents during task execution and result synthesis.

For this purpose, LLM-powered multi-agent systems realize an interaction layer \cite{bass2003software}. Externally, this layer facilitates the interaction between the LLM and its contextual environment. This includes interfacing with external data sources, tools, models, and other software systems or applications. These external entities can either generate or modify multi-modal artefacts or initiate further external processes. Internally, the interaction layer allows for organizing the task-management activity by providing a workspace for the collaboration between the LLM-powered agents. Thereby, LLM-powered multi-agent systems are characterized by diverse architectures implementing various architectural design options.

One of the central challenges for the effective operation of LLM-powered multi-agent architectures (as with many AI systems) lies in finding the optimal \textit{balance between autonomy and alignment} \cite{yudkowsky2016ai,bostrom2017superintelligence,russell2022artificial,wolf2023fundamental,hong2023metagpt}. On the one hand, the systems should be aligned to the goals and intentions of human users; on the other hand, the systems should accomplish the user-prompted goal in a self-organizing manner. 
However, a system with high autonomy may handle complex tasks efficiently, but risks straying from its intended purpose if not sufficiently aligned, resulting in unexpected consequences and uncontrollable side effects. 
Conversely, a highly aligned system may adhere closely to its intended purpose but may lack the flexibility and initiative to respond adequately to novel situations. 
Current systems exhibit diverse approaches and mechanisms to intertwine these \textit{cross-cutting concerns} \cite{kiczales1997aspect} throughout their architectural infrastructure and dynamics.

However, existing taxonomies and analysis frameworks for autonomous systems \cite{brustoloni1991autonomous,wooldridge1995intelligent,maes1995artificial,franklin1996agent,tosic2004towards} and multi-agent systems \cite{bird1993toward,dudek1996taxonomy,van2004design,moya2007towards} fall short in providing means to categorize and understand these challenges and involved architectural complexities posed by LLM-powered multi-agent systems.

This paper aims to bridge this gap by introducing a systematic approach in terms of a comprehensive multi-dimensional taxonomy. This taxonomy is engineered to analyze and classify how autonomous LLM-powered multi-agent systems balance the interplay between autonomy and alignment across different architectural viewpoints, encompassing their inherent aspects and mechanisms.

\begin{figure}[htb]
	\centering
	\includegraphics[width= 0.7 \textwidth]{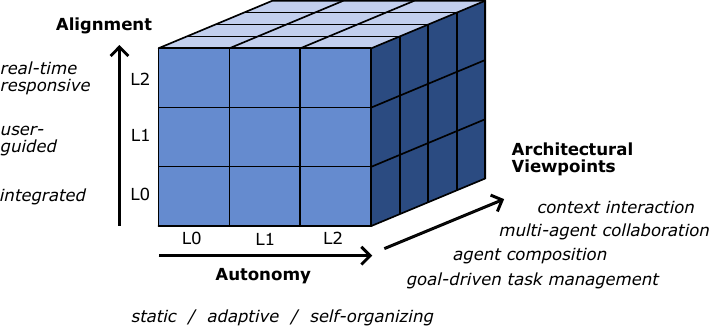}
	\caption{A simplified representation of the proposed multi-dimensional taxonomy for autonomous LLM-powered multi-agent systems. The x-axis represents the level of autonomy, the y-axis the level of alignment, and the z-axis the four applied architectural viewpoints. Characteristics of LLM-powered multi-agent systems can be assessed and classified by locating them within this taxonomic structure. For an in-depth discussion of the taxonomy, refer to Section \ref{sec:taxonomy}.}
	\label{figure:cube}
\end{figure}

The proposed taxonomy is built on the identification and specification of  architectural characteristics of such systems. It aims at understanding the complexities arising from the interplay of interdependent architectural aspects and mechanisms, each characterized by distinct levels of autonomy and alignment.
A simplified overview of the dimensions and levels applied in our taxonomy is represented by the cuboid shown in Fig.\ \ref{figure:cube}.

First, the synergy between autonomy and alignment manifests as a two-dimensional matrix with multiple hierarchical levels. This matrix captures a spectrum of nine distinct system configurations, ranging from systems that strictly adhere to predefined mechanisms (\textit{rule-driven automation}, L0/L0) to those that dynamically adapt in real-time, guided by evolving conditions and user feedback (\textit{user-responsive autonomy}, L2/L2).

Second, these configuration options are not imposed to the LLM-powered multi-agent system flatly. Instead, they are applied to multiple distinct architectural viewpoints \cite{kruchten19954+}, such as the system's functionality (\textit{goal-driven task management}), its internal structure (\textit{agent composition}), its dynamic interactions (\textit{multi-agent collaboration}) as well as the involvement of contextual resources such as tools and data (\textit{context interaction}). Stemming from these four viewpoints, we have discerned 12 architectural aspects, each with distinct autonomy and alignment levels. When integrated into our taxonomic system, they culminate in 108 single configuration options available for further combination. 
This granularity facilitates a nuanced analysis and assessment of the system's architectural dynamics resulting from the interplay between autonomy and alignment across the system architecture, laying the foundations for further analysis and reasoning about design decisions.

The contributions of this paper can be categorized as follows:
\begin{itemize}

    \item [(1)] \textbf{Architecture Specification:} We outline the architectural characteristics of autonomous LLM-powered multi-agent systems and propose a domain-ontology model represented as a UML class diagram structuring the architectural concepts and their interrelations relevant for our taxonomy. 

    \item [(2)] \textbf{Multi-dimensional Taxonomy:} We introduce a comprehensive multi-dimensional taxonomy tailored to analyze and understand how autonomous LLM-powered multi-agent architectures balance the dynamic interplay between autonomy and alignment across different architectural viewpoints. For this purpose, our taxonomy provides hierarchical levels for both autonomy and alignment, which are applied to system viewpoints such as goal-driven task management, agent composition, multi-agent collaboration, and context interaction, thus incorporating a third dimension. Level criteria for autonomy and alignment are specified for several architectural aspects inherent to these viewpoints.

    \item [(3)] \textbf{Taxonomic Classification of Selected Systems:} We demonstrate the applicability and effectiveness of our taxonomy by assessing and comparing a selection of seven representative autonomous LLM-powered multi-agent systems. This taxonomic classification provides insights into the architectural dynamics and balancing strategies of the analyzed systems. Moreover, it identifies challenges and development potentials, not only concerning the interplay between autonomy and alignment. The application of the taxonomy also serves as a first empirical validation.
    
\end{itemize}

Through these contributions, we aim to provide a systematic framework for analyzing, comparing, and understanding the architectural dynamics and complexities of LLM-powered multi-agent systems, thus contributing to the ongoing efforts towards building more reliable and efficient multi-agent systems.

\textbf{Structure of the Paper.}
The remainder of this paper is structured as follows.
Section \ref{sec:background} discusses related work on taxonomies for intelligent systems and gives an overview of existing autonomous LLM-based multi-agent systems.
Section \ref{sec:architecture} outlines the key characteristics and specifies foundational concepts of autonomous LLM-powered multi-agent architectures relevant for the taxonomy.
In Section \ref{sec:taxonomy}, we introduce our multi-dimensional taxonomy, which incorporates specifications of autonomy and alignment levels and their application to architectural viewpoints and aspects of LLM-powered multi-agent systems.
Section \ref{sec:comparison} showcases the utility of our taxonomy, as we analyze and categorize selected current multi-agent systems.
In Section \ref{sec:discussion}, we discuss the insights gained from this analysis.
Finally, Section \ref{sec:conclusion} concludes the paper.

\section{Background and Related Work}
\label{sec:background}

In this section, we discuss the application of taxonomies for autonomous agents and multi-agent systems (Section \ref{subsec:bg-taxonomies}) and give an overview of the state-of-the-art of LLM-powered multi-agent systems (Section \ref{subsec:bg-architectures}).

\subsection{Taxonomies and Intelligent Systems}
\label{subsec:bg-taxonomies}

Taxonomies represent structured classification schemes employed  to categorize objects in a hierarchical manner according to specific criteria. They are a popular means for structuring, measuring or comparing various kinds of approaches such as methods, techniques or technologies. They find applications in a wide range of disciplines and domains such as software engineering \cite{usman2017taxonomies} or explainable artificial intelligence (XAI) \cite{arrieta2020explainable}.

The field of intelligent and autonomous systems spans a variety of configurations and operational structures, with some systems operating as individual entities and others involving multiple interacting agents. Reflecting this variety, the taxonomies proposed over the years have largely followed two main trajectories: those focusing on autonomous systems and those focusing on multi-agent systems.

\textbf{Taxonomies for Autonomous Systems} mainly categorize systems based on the level and type of autonomy, intelligence, learning capabilities, and ability to interact with their environment. 
These taxonomies, such as those by \cite{wooldridge1995intelligent,brustoloni1991autonomous,maes1995artificial,franklin1996agent,tosic2004towards}, are essential for understanding the spectrum of capabilities and complexities inherent to these systems.
In particular, Wooldridge and Jennings \cite{wooldridge1995intelligent} present a comprehensive taxonomy that classifies intelligent agents based on key properties such as autonomy, social ability, reactivity, and proactiveness. Their classification sheds light on the independent operational capabilities of single-agent systems. In addition, Brustoloni \cite{brustoloni1991autonomous} introduces a taxonomy centered around the idea of autonomy levels, drawing the distinction between autonomous, semi-autonomous, and non-autonomous systems. This provides a valuable lens through which the extent of an agent's independence can be analyzed.
Maes' taxonomy \cite{maes1995artificial} focuses on situating agents within a landscape defined by their reasoning and learning capabilities. The work provides a robust framework for assessing the cognitive dimensions of an agent, ranging from reflex agents to learning agents. Franklin and Graesser \cite{franklin1996agent}, in their work, delve into the interaction between autonomous agents and their environment, leading to a taxonomy that is heavily contextual and environmental-dependent.
Lastly, Tosic and Agha \cite{tosic2004towards} put forth a taxonomy that embraces the diversity in the field, offering a multi-faceted perspective on autonomous agents, taking into consideration their design, behavior, and interaction capabilities.
Besides these scientific frameworks, further industry-focused and more pragmatic taxonomies are in use, such as the taxonomy provided by SAE international \cite{sae2016taxonomy} serving as a foundational standard for self-driving cars with definitions for levels of driving automation; from no driving automation to full automation.

However, while these taxonomies offer valuable insights into the capabilities and behaviors of autonomous systems, they don't inherently address the complexity and nuances involved when multiple agents powered by large language models are working together within a multi-agent architecture. Hence, the need for a dedicated taxonomy for such systems is evident.

\textbf{Taxonomies for Multi-Agent Systems}, on the other hand, extend beyond the confines of individual agent characteristics, integrating the dynamics of interactions and collaborations among multiple agents. 
The landscape of multi-agent systems taxonomies provides various works focusing on different aspects of these complex systems \cite{bird1993toward,dudek1996taxonomy,van2004design,moya2007towards}. In particular, Bird et al.\ developed a taxonomy rooted in the communication and cooperation strategies among agents, investigating crucial factors such as communication methods, task decomposition, resource sharing, and conflict resolution \cite{bird1993toward}.
Similarly, Dudek et al.\ offered a taxonomy specifically for multi-robot systems \cite{dudek1996taxonomy}. This taxonomy, while primarily focusing on robotic applications, can be generalized to other multi-agent systems, considering important aspects like team size, communication topology, team organization, and team composition.
In a different vein, Van Dyke Parunak et al.\ proposed a taxonomy for distributed AI systems, putting emphasis on environmental aspects and interaction modalities, thus highlighting the importance of the agents' ability to interact with and manipulate their environment \cite{van2004design}.
Further broadening the field, Moya et al.\ proposed a comprehensive taxonomy for multi-agent systems based on characteristics such as the nature of the agents, the environment in which they operate, the communication protocols they use, and the tasks they perform \cite{moya2007towards}. The taxonomy also thoroughly examined the various types of interactions among the agents, including cooperation, coordination, and negotiation.

While these taxonomies have contributed significantly to our understanding of communication protocols and agent constellation within multi-agent systems, they were developed prior to the advent of large language models (LLMs), and thus do not encapsulate the characteristic challenges associated with LLM-based multi-agent architectures. 
In this context, autonomous agents, also as members of multi-agent networks, often have been used as a kind of metaphor \cite{franklin1996agent} for intelligent and interacting components following rule-based communication protocols and bundling a set of specific skills to interact with their environment (e.g., equipped with certain sensors and actuators, cf.\ multi-robot systems \cite{dudek1996taxonomy}). 
LLMs have introduced a new degree of reasoning capabilities, enabling the creation of genuinely intelligent agents operating within collaborative networks in an autonomous manner.

Moreover, while the concepts of autonomy and alignment are often discussed in AI literature \cite{narendra2012stable,russell2019human} and also the system's architecture plays a fundamental role in software engineering \cite{bass2003software}, none of the existing taxonomies for autonomous systems or for multi-agent systems has so far applied a systematic approach to either investigate architectural aspects or combine the concepts of autonomy and alignment for analyzing the systems.

In the light of these limitations, our work seeks to develop a new taxonomy specifically tailored to LLM-powered multi-agent systems. Our aim is to provide a taxonomy that captures the unique aspects and challenges of LLM-powered architectures, especially with regard to how autonomy and alignment are balanced across architectural aspects and viewpoints, offering a systematic framework for understanding and designing these complex systems.

\subsection{Current LLM-based Agent Systems}
\label{subsec:bg-architectures}

The advent and widespread use of large language models (LLMs) like GPT-3 \cite{brown2020language} have opened up new opportunities for creating increasingly sophisticated and human-like AI systems. These models empower the design of intelligent agents with advanced capabilities to comprehend and generate human-like text, thus enriching the interaction and experience for end-users.
However, the application of LLMs also brings forth several challenges, as highlighted by \cite{kaddour2023challenges}. Among these, one main challenge lies in handling task complexity, particularly when dealing with intricate tasks that necessitate a well-coordinated execution of numerous interconnected sub-tasks \cite{shen2023hugginggpt} and the interaction with further tools and data.
In response to this, autonomous multi-agent systems utilizing the reasoning abilities of LLMs have emerged. These systems address complexity by intelligently breaking down larger goals into manageable tasks, which are then accomplished by multiple collaborating agents specializing in a specific role and equipped with distinct competencies. 
For this purpose, these systems realize an \textit{interaction layer} \cite{bass2003software} providing a workspace for multiple collaborating agents, each connected with an LLM. The reasoning competencies are enhanced, as needed, by the agent's access to contextual resources, such as specific expert tools, data sets, further foundation models, and external applications, which allow the agents to gain information from and to impact their environment by creating or modifying multi-modal artefacts or by triggering external processes. The agents collaborate, bringing their capabilities to bear on the problem, and their results are subsequently combined to achieve the overall goal. 

Currently, several projects are established that aim at realizing such autonomous AI architectures for accomplishing complex tasks based on multiple interacting agents and powered by large language models (LLMs). 
Exemplary but representative autonomous multi-agent systems are \textsc{AutoGPT} \cite{autogpt2023}, \textsc{BabyAGI} \cite{babyagi2023}, \textsc{SuperAGI} \cite{superagi2023}, \textsc{HuggingGPT} \cite{shen2023hugginggpt}, \textsc{CAMEL} \cite{li2023camel}, \textsc{AgentGPT} \cite{agentgpt2023} and \textsc{MetaGPT} \cite{hong2023metagpt}. For a categorization and comparison of these selected system architectures using the developed taxonomy, see Section \ref{sec:comparison}.
Among these systems, we can distinguish those providing general-purpose task management and problem solving with generic agent types and collaboration mechanics \cite{autogpt2023,babyagi2023,superagi2023,shen2023hugginggpt} and those systems designed for specific application domains with corresponding domain agents and processes, such as for the domain of software development \cite{hong2023metagpt,li2023camel,qian2023communicative}.

Some of these recent multi-agent systems as well as further related projects such as \textsc{Gorilla} \cite{patil2023gorilla} or \textsc{Voyager} \cite{wang2023voyager} are built upon the \textsc{LangChain} Python framework \cite{langchain2022}, which allows to realize the aforementioned interaction layer to define agents and chains of tasks as well as the access to and interplay between large language models and contextual resources in terms of data resources (such as vector databases \cite{johnson2019billion}) or various expert tools. For this purpose, predefined components such as agent types and prompt templates, such as for data interaction, can be reused.

Besides these open-source software projects, further scientific projects and approaches are identifiable that leverage multiple agents or personas powered by LLMs for task management and problem solving \cite{wang2023unleashing,hao2023chatllm}.
Moreover, the interplay between multiple LLM-powered agents is also addressed in other related contexts, such as for simulating the interaction between multiple personas or roles \cite{park2023generative,gao2023s}, especially with focus on debating and thereby addressing challenges of hallucinations \cite{du2023improving} or the \textit{degeneration-of-thought (DoT)} \cite{liang2023encouraging}, or for developing conversations and behavior provided by non-playable characters (NPCs) in role-playing video games \cite{csepregi2023effect}. 

A recent survey of LLM-powered autonomous agents is provided by \cite{wang2023survey}, which focuses on investigating and comparing the agents' characteristics and capabilities in terms of profile generation, memory operations and structures, planning, tool integration and learning strategies. Complementing this, another recent survey \cite{xi2023rise} offers an expansive overview of existing approaches, contextualizing them with foundational technical, methodical, and conceptual paradigms formative for LLM-powered multi-agent systems. 

However, as we dive into the specifics of current autonomous LLM-powered multi-agent systems, striking the right balance between autonomy and alignment emerges as a central challenge. These AI systems must navigate a fine line – being autonomous enough to organize the interplay between multiple LLM-powered agents and contextual resources to accomplish complex tasks consisting of various interconnected sub-tasks, but also adequately aligned to the intentions and goals of users. This especially proves challenging, since the specified and prompted user goal might not exactly represent the user's intentions \cite{boehm1988understanding}, resulting in unexpected consequences and uncontrollable side effects.

Given the exploratory state of the field, current systems exhibit a wide range of architectures, each with its unique blend of autonomy and alignment dispersed across various architectural components and mechanisms. 
The diversity in these systems illuminates the different strategies and designs adopted to address this balancing. However, it also signifies the lack of a systematic approach, underscoring the importance of a taxonomy that can provide a structured understanding and comparison of these systems.

\section{Architecture Specification}
\label{sec:architecture}
In this section, we specify architectural foundations relevant for our taxonomy. Section \ref{subsec:aa-characteristics} provides an overview of architectural and behavioral characteristics of autonomous LLM-powered multi-agent systems. Following this, we delve deeper into the architectural key concepts and their interrelationships through a domain-ontology model (see Section \ref{subsec:aa-ontology}). 

\begin{figure}[htb]
	\centering
	\includegraphics[width= 0.9\textwidth]{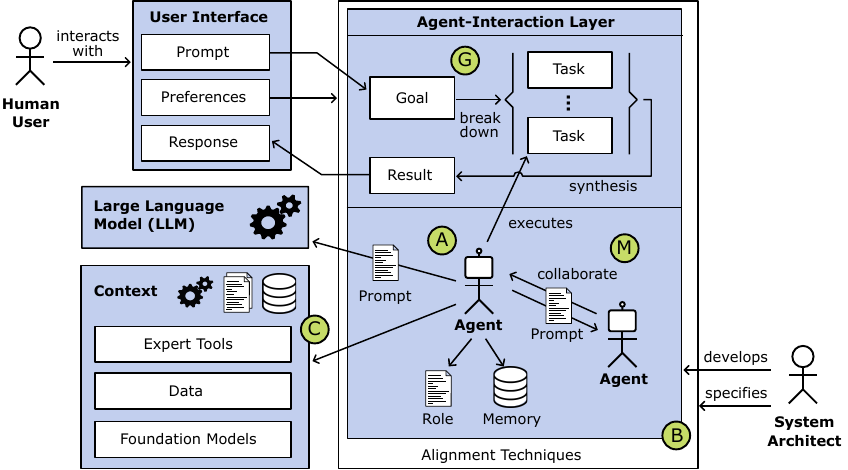}
	\caption{Overview of the primary characteristics of autonomous multi-agent systems powered by large language models (LLMs) and enhanced by contextual resources like tools and data. A description of these characteristics is provided in Section \ref{subsec:aa-characteristics}, and for an in-depth exploration of architectural concepts, refer to Section \ref{subsec:aa-ontology}.
 }
	\label{figure:architecture}
\end{figure}

\subsection{Characteristics Overview}
\label{subsec:aa-characteristics}

In the following, we outline the main architectural characteristics of autonomous LLM-powered multi-agent systems, as illustrated in Fig.\ \ref{figure:architecture}.
\begin{itemize}

    \item [\circled{G}] \textbf{Goal-driven Task Management.} 
Autonomous LLM-powered multi-agent systems are designed to accomplish user-prompted goals or complex tasks. For this purpose, the system employs an interactive and multi-perspective strategy for problem solving, often referred to as deep reasoning or \textit{slow thinking} \cite{kahneman2011thinking} enabled by the capabilities of large language models (LLMs) and the advantages of contextual resources. When faced with such challenges, the system adeptly breaks down the complex task into smaller, manageable tasks. These sub-tasks are subsequently distributed among various agents, each equipped with specific competencies. 
A crucial aspect of this \textit{divide \& conquer} strategy lies in the effective orchestration of these interconnected sub-tasks and the subsequent synthesis of partial results, ensuring a seamless and cohesive final result.

    \item [\circled{A}] \textbf{LLM-Powered Intelligent Agents.} 
At the core of these systems, intelligent agents structure the system as the foundational components. Each agent is endowed with a unique set of competencies, which include a clearly defined role, an individual memory, as well as access to further contextual resources, such as data, tools, or foundation models (see below), required for solving the tasks assigned to them. 
The backbone of their reasoning and interpretative capabilities is rooted in the incorporation of large language models (LLMs). This enables the agents not only to reflect upon the tasks or to plan and process the assigned tasks efficiently, but also to access and utilize contextual resources, as well as to communicate with other agents.

    \item [\circled{M}] \textbf{Multi-Agent Collaboration.} 
The interaction layer provides the workspace for a network of such collaborating LLM-powered agents. 
While executing the assigned tasks, these specialized agents collaborate with each other via prompt-driven message exchanges to delegate responsibilities, seek assistance, or evaluate the results of tasks undertaken by their peers.
Key to the agents' collaboration is to effectively combine the strengths of each agent to collectively meet the defined goals, exemplifying \textit{cognitive synergy}. While individual skills are important, the power of these systems emerges from the coordinated efforts of the collective, a concept articulated by Minsky in his idea of the \textit{society of mind} \cite{minsky1988society}. 

    \item [\circled{C}] \textbf{Context Interaction.} 
Some tasks require the utilization of contextual resources, such as expert tools, data, further specialized foundation models, or other applications. These resources extend their ability to gather environmental information, create or modify artefacts, or initiate external processes. 
Leveraging these resources enables the agents to better understand and respond to their operational context and to effectively execute complex tasks.
This capacity for contextual adaptation, augmented by the integration of various resources, contributes to a more versatile and comprehensive system that can address diverse challenges and requirements.

    \item [\circled{B}] \textbf{Balancing Autonomy and Alignment.} 
The dynamics of LLM-powered multi-agent systems are characterized by a complex interplay between autonomy and alignment. As captured in Fig.\ \ref{figure:triadic-relationship}, this complexity can be traced back to the triadic interplay and inherent tensions among the primary \textit{decision-making entities}: human users, LLM-powered agents, and governing mechanisms or rules integrated into the system.
\textit{Alignment}, in this context, ensures that the system's actions are in sync with human intentions and values. On the other side of the spectrum, \textit{autonomy} denotes the agents' inherent capacity for self-organized strategy and operation, allowing them to function independent of predefined \textit{rules and mechanism} and without human \textit{supervision}.
Moreover, in systems steered by user-prompted goals, it becomes pivotal to differentiate between generic alignment aspects, in terms of \textit{mechanisms} predefined by system architects to inform core functionalities, and user-specific preferences \textit{customized} by the system users themselves.

\begin{figure}[htb]
	\centering
	\includegraphics[width= 0.5\textwidth]{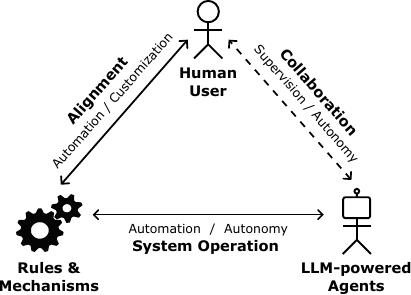}
	\caption{Triadic interplay and dynamic tensions between the decision-making entities in LLM-powered multi-agent systems.
 }
	\label{figure:triadic-relationship}
\end{figure}

However, from an architectural perspective, autonomy and alignment transform into \textit{cross-cutting concerns} \cite{kiczales1997aspect}. They traverse components and mechanisms across the entirety of the system's architecture, influencing the communication between agents, the interaction with contextual resources, and more.
Achieving a balanced configuration of autonomy and alignment is a crucial challenge, which directly impacts the system's efficiency and effectiveness.

\end{itemize}

In the following Section \ref{subsec:aa-ontology}, we elaborate on the architectural details of these characteristics.

\begin{figure}[htb]
	\centering
	\includegraphics[width= \textwidth]{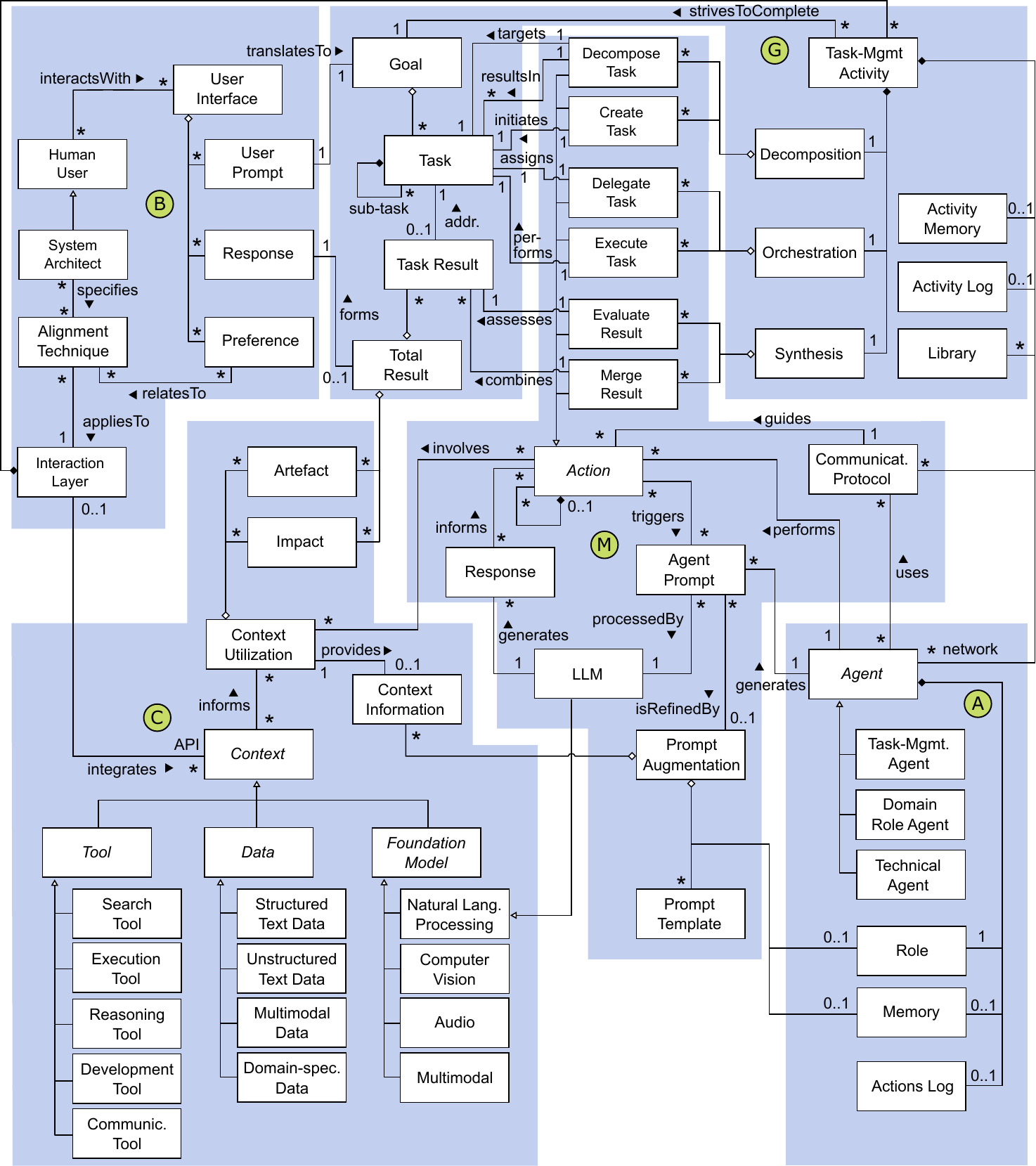}
	\caption{Domain-ontology model represented as UML class diagram structuring selected architectural concepts and concept relations relevant for the domain of autonomous LLM-powered multi-agent systems. For a comprehensive exploration of this model, refer to Section \ref{subsec:aa-ontology}.}
	\label{figure:ontology}
\end{figure}

\subsection{Specification of Architectural Components}
\label{subsec:aa-ontology}

Fig.\ \ref{figure:ontology} illustrates a structured overview of selected concepts and their interrelations relevant for the addressed domain of autonomous LLM-powered multi-agent systems in terms of a domain-ontology model. 
Domain ontologies, embraced across fields from philosophy to information systems, facilitate a shared understanding of domain-specific concepts \cite{sowa1995top}. While they aid automated knowledge dissemination among software entities as \textit{formal ontologies} \cite{gruber1995toward}, they are also devised as \textit{conceptual models} to support human understanding of the addressed domain \cite{kitchenham1999towards,guizzardi2002general,haendler2020ontology}.

Our domain ontology is represented as a conceptual model in terms of a class diagram of the \textit{Unified Modeling Language (UML2)} \cite{UML}, which allows for organizing the identified concepts as classes and their relationships in terms of generalizations and kinds of associations with indicated multiplicities (i.e.\ amount of objects involved in a relationship). For further details on the applied diagram notations, the UML specification serves as a comprehensive guide \cite{UML}.

The primary objective of the presented model in Fig.\ \ref{figure:ontology} is to mirror architectural concepts especially relevant to our taxonomy's scope. In doing so, it deliberately adopts a high-level view, abstracting from technical details and specifics typical of individual systems to support clarity and accessibility. 
For example, while diving into the complexities of the agent's memory usage, as for reflecting and combining task instructions or for planning steps and actions, is undoubtedly worthy of thorough exploration \cite{park2023generative,wang2023plan}, it falls outside the narrow scope of our taxonomy.
This approach ensures that actual multi-agent systems can be regarded as potential instances of this conceptual framework. It's worth noting that this doesn't preclude the addition of more specific technical components and mechanisms as systems evolve.

The domain-ontology model derives from an examination of the code and architectural documentation of several representative multi-agent architectures, especially \textsc{AutoGPT} \cite{autogpt2023}, \textsc{SuperAGI} \cite{superagi2023}, and \textsc{MetaGPT} \cite{hong2023metagpt}, the \textit{Generative Agents} project \cite{park2023generative}, as well as the \textsc{LangChain} framework \cite{langchain2022}. The latter serves as the foundational infrastructure for some of the assessed multi-agent systems (refer to Section \ref{subsec:bg-architectures}). Through an iterative process, we analyzed these systems and frameworks to understand their components, interactions, and overarching structures. This analysis facilitated the identification and abstraction of recurrent architectural characteristics prevalent among these architectures.

The concepts of the model are arranged into thematic blocks corresponding to the system characteristics identified in Section \ref{subsec:aa-characteristics}, such as \circled{G}. In the following, we delve into these concepts and their main interrelations. Further details are provided in the domain-ontology diagram illustrated in Fig.\ \ref{figure:ontology}.

\circled{G} \textbf{Concepts of Goal-driven Task Management.}
Typically, a \texttt{Human User} initiates system operations via a \texttt{User Prompt} through the \texttt{User Interface}. Most of these systems employ \textit{single-turn prompting} to convey intricate \texttt{Goals} \cite{santu2023teler}. The prompts can be enriched with detailed instructions, exemplifications like reasoning sequences \cite{wei2022chain}, role specifications, or output expectations \cite{kaddour2023challenges}. Systems may also permit the definition of \texttt{Preferences} to better align AI operations with user objectives. Besides textual user input, speech, images, videos, or mode combinations are conceivable, for example. 
Internally, this user-prompted \texttt{Goal} (which might represent a directive, problem, question, or mission) undergoes decomposition into \texttt{Tasks} or \texttt{Sub-Tasks} to be manageable by the  \texttt{Agents}. These tasks can be interconnected in different ways, such as \textit{sequential tasks} or \textit{graph tasks} \cite{shen2023hugginggpt}, which requires appropriate task prioritization.
Task decomposition is the first of three core phases within the \texttt{Task-Management Activity}: 
\begin{itemize}
    \item \texttt{Decomposition}: Breaking down complex tasks into manageable \texttt{Tasks} and \texttt{Sub-Tasks}; optionally resolving dependencies between them, resulting in a prioritized list of \texttt{Tasks}. 
    \item \texttt{Orchestration}: Organizing the distribution and delegation of \texttt{Tasks} among suitable \texttt{Agents}.
    \item \texttt{Synthesis}: Evaluating and combining \texttt{Task Results} as well as finally presenting a unified \texttt{Total Result}.
\end{itemize}
Furthermore, each \texttt{Task-Management Activity} embodies an \texttt{Activity Log} and an \texttt{Activity Memory}.  To maintain transparency and traceability of all \texttt{Actions} performed, an \texttt{Activity Log} captures all relevant action details throughout an activity, while the \texttt{Activity Memory} distills and retains key insights. In addition, systems might feature a \texttt{Library}, a repository storing best practices, lessons learned, or reusable knowledge, such as \texttt{Prompt Templates} \cite{white2023prompt} or specific information like API credentials.
\texttt{Actions} within this activity are delegated to specialized \texttt{Agents}—each characterized by a distinct \texttt{Role}, \texttt{Type}, and further competencies. Essential for the actions' success is their interaction with various kinds \texttt{Context} —ranging from \texttt{Data} and expert \texttt{Tools} to foundational \texttt{Models} (detailed below).
Once all partial tasks are completed, the \texttt{Task Results} are integrated and combined into a \texttt{Total Result} addressing the prompted \texttt{Goal}. This result might also include multiple \texttt{Artefacts}, encompassing text, graphics, multimedia, and more. The nature and involvement of tools applied, such as \texttt{Execution} or \texttt{Development Tools}, can lead to varied \texttt{Impacts}, such as triggering external processes. Finally, the \texttt{Response}, a summation of the result, is relayed to the user through the user interface.

\circled{A} \textbf{Concepts of LLM-Powered Intelligent Agents.}
Within each \texttt{Task-Management Activity}, a set of intelligent \texttt{Agents} collaborate, forming a multi-agent \texttt{Network}. These agents derive their advanced reasoning capabilities from \texttt{Large Language Models} (LLMs) \cite{brown2020language,kaddour2023challenges,naveed2023comprehensive}, which are involved in performing different kinds of \texttt{Actions}, each related to a certain \texttt{Task} and/or contributing to its \texttt{Task Result}. Each agent is differentiated by its unique \texttt{Role} in the activity and possesses an individual \texttt{Memory}—a repository that encompasses condensed experiences and knowledge gained by the agent. It can manifest in multiple formats—be it textual records, structured databases, or embeddings. Oriented to human memorization, in multi-agent systems, we also see a combination of short-term memory (i.e., compressed information transmission via the context window) and approaches for long-term memory \cite{wang2023augmenting,wang2023recursively}, such as by leveraging vector databases (see below).
The encapsulated history of the agent's actions might also be chronicled in an \texttt{Actions Log}.
Furthermore, different generic \texttt{Agent} types can be distinguished in terms of their roles, and their unique functionalities within the collaborative agent network.
\begin{itemize}
    \item \texttt{Task-Management Agents}: These agents are specialized in organizing the processes related to the task-management activity \cite{autogpt2023,babyagi2023,li2023camel}. 
    \begin{itemize}
        \item \texttt{Task-Creation Agent}: Generating new tasks, which optionally also includes deriving tasks by breaking down complex tasks.
        \item \texttt{Task-Prioritization Agent}: Assigning urgency or importance to tasks, which includes to resolve the dependencies between the tasks.
        \item \texttt{Task-Execution Agent}: Ensuring efficient task completion.
    \end{itemize}
    \item \texttt{Domain Role Agents}: These agents are domain-specific experts. They excel in specialized roles within the application domain \cite{park2023generative}, collaborating with peer role agents when needed. Examples encompass roles in the software-development process, such as project manager, software architect, developer, or QA engineer \cite{hong2023metagpt,li2023camel}.
    \item \texttt{Technical Agents}: These agents are tech-savvies, typically tasked with interfacing with technical platforms or development tools. Exemplary technical agents are represented by the \texttt{SQL Agent} for database interactions or the \texttt{Python Agent} for developing Python scripts \cite{langchain2022,hong2023metagpt}.
\end{itemize}
An essential distinction to note is the variability in agent memory reliance. While some agents harness the power of memory or an action log, e.g., for reflecting or planning tasks, others function devoid of these recollections. Specifically, for technical aspects or actions that demand an unprejudiced or unbiased lens, agents without memories are often preferred.

\circled{M} \textbf{Concepts of Multi-Agent Collaboration.}
As detailed above, each \texttt{Task-Management Activity} involves a set of multiple collaborating \texttt{Agents} with different roles and competencies as well as driven by the reasoning capabilities of utilized large language models (LLMs). This reasoning power enables the agents to reflect, plan and process the assigned tasks \cite{wang2023plan,park2023generative} as well as to interact with other agents \cite{hong2023metagpt}.
In particular, the \texttt{Agents} execute different kinds of \texttt{Actions} which in sum aim at achieving the user-prompted \texttt{Goal}. 
In particular, the following sub-types of \texttt{Action} performed by the \texttt{Agents} can be distinguished: 
\begin{itemize}
    \item \texttt{DecomposeTask}: Breaking down a task into multiple sub-tasks, optionally ordering and prioritizing the tasks.
    \item \texttt{Create Task}: Defining and generating new tasks.
    \item \texttt{DelegateTask}: Delegating a task to another agent, addressed as \texttt{Receiver}.
    \item \texttt{ExecuteTask}: Actually executing a given task.
    \item \texttt{EvaluateResult}: Assessing the outcomes of a task.
    \item \texttt{MergeResult}: Integrating or combining two or more task results.
\end{itemize}
Thereby, each \texttt{Action} can be part of another \texttt{Action}, which are, however, performed in the context of a certain phase of the \texttt{Task-Management Activity} (see Fig.\ \ref{figure:ontology}). 
Furthermore, each \texttt{Action} can include multiple interactions with an LLM. The LLM's reasoning capabilities are employed in multiple directions within an \texttt{Action}, such as for reflecting memories and instructions, observing existing results, planning steps and/or weighing options to proceed \cite{wang2023plan,park2023generative,hong2023metagpt}.
For this purpose, an \texttt{Agent Prompt} generated by an \texttt{Agent} and triggered within a certain \texttt{Action} is send to and then processed by the LLM, which generates a \texttt{Response} informing and/or guiding the next steps within the triggering action. An action might also include \texttt{Context Utilization}.
Before the LLM receives the \texttt{Agent Prompt}, it may undergo \texttt{Prompt Augmentation} \cite{shum2023automatic}. 
This process can integrate additional specifics like the aspects or parts of the agent's \texttt{Role} or \texttt{Memory}, \texttt{Context Information} (e.g., data excerpts) acquired from previous \texttt{Context Utilization}, or chosen \texttt{Prompt Templates} prepared and/or adapted for certain kinds of actions \cite{langchain2022}. Such agent-driven \textit{prompt engineering} is pivotal for LLM-powered multi-agent systems.

Direct collaborations involving two or more agents typically rely on prompt-driven communication sequences or cycles. For instance, a \texttt{Delegate Task} action directed at a \texttt{receiver} agent might convey information, place a request, initiate a query, or suggest a potential course of action. Subsequently, the \texttt{Evaluate Result} action provides feedback by validating or refuting, and agreeing or disagreeing with the presented results \cite{wooldridge2009introduction}.
A \texttt{Communication Protocol} provides a structured framework and methodology for agents' collaboration, guiding the execution of specific \texttt{Actions} by establishing rules and mechanisms for message exchanges within the multi-agent network \cite{singh1998agent,labrou1998semantics,wooldridge2009introduction}.
For instance, in LLM-powered multi-agent systems, the following distinct protocols are observable, each built upon the basic mechanisms of an interplay between instruction and execution, with optional subsequent result evaluation:
\begin{itemize}
    \item \textit{Strict finite processes} or execution chains with predefined action sequences, interactions between predefined agents, and typically having a well-defined endpoint, which might represent the production of a specific output or artefact \cite{hong2023metagpt}.
    \item \textit{Dialogue cycles} characterized by alternating \texttt{DelegateTask} and \texttt{ExecuteTask} actions between two agents, creating a feedback loop of instruction and execution \cite{li2023camel}.
    \item \textit{Multi-cycle process frameworks} with interactions between generic agent types, allowing for greater dynamism in agent interactions \cite{autogpt2023,babyagi2023}. 
\end{itemize}
In all these exemplary cases, dedicated \texttt{Agent Types} are defined and coupled with the corresponding types of \texttt{Action}. Further details are discussed in Section \ref{subsec:comp-interpretation}.

\circled{C} \textbf{Concepts of Context Interaction.}
For executing the task-related actions, the LLM-powered agents are able to leverage specialized competencies and further information provided by additional \texttt{Context} which can be distinguished into \texttt{Tools}, \texttt{Data}, and \texttt{Foundation Models} \cite{shen2023hugginggpt} (see Fig.\ \ref{figure:ontology}). 

\begin{itemize}
\item \texttt{Tools} in terms of contextual resources for multi-agent systems can be categorized into the following distinct groups:
    \begin{itemize}
    \item \texttt{Search and Analysis Tools}: These tools offer specialized capabilities for probing and analyzing data, allowing agents to derive insights from vast information pools efficiently; such as search engines for the web. 
    \item \texttt{Execution Tools}: These are responsible for interfacing with and executing tasks within other environments, like software applications, ensuring seamless operation across platforms. 
    \item \texttt{Reasoning Tools}: Enhancing the capacity for logical thought, these tools bolster reasoning capabilities in specialized areas such as computational intelligence. For instance, platforms like \textsc{Wolfram Alpha} empower agents with advanced computational skills.
    \item \texttt{Development Tools}: Tailored for software development endeavors, these tools streamline the process of coding, debugging, and deploying solutions within the multi-agent framework. 
    \item \texttt{Communication Tools}: These facilitate interactions with external entities by supporting functionalities like sending and receiving emails, ensuring agents can effectively communicate outside their native environments.
    \end{itemize}

    \item \texttt{Data} types in multi-agent architectures encompass:
    \begin{itemize}
    \item \texttt{Structured Text Data}: This refers to data that adheres to a defined model or schema, such as data found in traditional relational databases. It offers predictability and is easily queryable.
    \item \texttt{Unstructured Text Data}: This data lacks a pre-defined model. An example is content found within PDF documents. For optimal processing by LLMs, unstructured text is typically stored in vector databases like \textsc{Pinecone} or \textsc{Chroma}. These databases support semantic searches through vector embeddings, bridging the gap between structured and unstructured data \cite{mikolov2013efficient,johnson2019billion}.
    \item \texttt{Multimodal Data}: Beyond just text, this category encapsulates various formats including videos, pictures, and audio. Specialized tools are employed to extract textual information from these formats, making them amenable to processing by LLMs.
    \item \texttt{Domain-specific Data}: This data is tailored for particular sectors or areas of expertise. Examples include proprietary company data or external data sources specific to fields like law or medicine.
    \end{itemize}
    
    \item \texttt{Foundation Models} refer to expansive machine learning models trained on vast amounts of data. These models are versatile, suitable for addressing a variety of tasks across different modalities such as language, vision, and audio/speech, as well as combinations thereof \cite{bommasani2021opportunities}. Based on their modalities, we categorize them as follows:
    \begin{itemize}
    \item \texttt{Natural Language Processing (NLP) Models}: These focus primarily on understanding and generating human language. \texttt{LLMs} fall under this category of \texttt{NLP Models}. While there are general-purpose LLMs available, specialized models tailored to specific domains and tasks also exist, having been trained on corresponding niche data sets.
    \item \texttt{Computer Vision Models}: Aimed at processing and understanding images or videos.
    \item \texttt{Audio Models}: Specialized in processing and interpreting audio signals, including speech.  
    \item \texttt{Multimodal Models}: Designed to handle multiple types of data simultaneously, combining aspects of \texttt{NLP}, \texttt{Vision}, and \texttt{Audio}.
\end{itemize}
The machine learning landscape shows a multitude of specialized foundation models, with Large Language Models (LLMs) standing out prominently \cite{zhao2023survey}. Platforms like \textsc{Hugging Face} even offer access to numerous models provided by the global machine learning community.

\end{itemize}
Access to LLMs, as well as associated resources such as tools, foundation models, and external data resources, is typically facilitated through Application Programming Interfaces (APIs) \cite{patil2023gorilla}. The access details for these APIs are integrated into the \texttt{Interaction Layer}. For instance, these details might be housed within a dedicated \texttt{Library} module for streamlined interfacing (see above).
Moreover, multiple of these contextual resources can be combined within a single \texttt{Action}. For example, a certain expert tool could employ a selected foundation model to analyze a given dataset.
Finally, \texttt{Context Utilization} might involve the creation or modification of \texttt{Artefacts}. Beyond mere artefact manipulation, this utilization can manifest as external \texttt{Impact}, such as initiating external processes in other software applications or triggering workflows that influence broader systems.

\circled{B} \textbf{Concepts of Balancing Autonomy and Alignment.}
Autonomy and alignment represent \textit{cross-cutting concerns} \cite{kiczales1997aspect}, influencing various architectural concepts and mechanisms. Nevertheless, they also distinctly materialize in specific concepts within our ontology model.
\textit{Alignment}, on the one hand, primarily manifests through the implementation of dedicated \texttt{Alignment Techniques} by the \texttt{System Architect} into the system architecture of the \texttt{Interaction Layer}. 
These techniques might include foundational infrastructural approaches or procedural controls for system components, framed as constraints, rules, or limitations.
Moreover, alignment can be expressed by the \texttt{System User}. The user-prompted \texttt{Goal} can be further refined pre-runtime through supplementary \texttt{Preferences} provided by the \texttt{Human User} via the \texttt{User Interface}. In addition, real-time adaptability can be offered by multi-agent systems, necessitating instantaneous system responsiveness. A more in-depth exploration of this is available in Section \ref{subsubsec:alignment}.
However, apart from the alignment achieved within the interaction layer by the multi-agent system, which our taxonomy addresses, there also exist alignment methodologies specifically tailored for the employed LLMs or foundation models  \cite{askell2021general,wolf2023fundamental}.
\textit{Autonomy}, on the other hand, primarily surfaces from the capability of the multi-agent system to fulfill the designated \texttt{Goal} autonomously through self-organized strategy and task execution.
Not only do individual collaborative intelligent \texttt{Agents} utilize the \texttt{LLM} for reflecting, planning, or performing reasoning-intensive actions pertinent to their roles, but the overarching organization of the \texttt{Task-Management Activity}, along with other infrastructural or dynamic elements, might also be directed in a self-organized manner, steered by LLM-powered \texttt{Agents}. Further details on this can be found in Section \ref{subsubsec:autonomy}.
Navigating the complex interplay between autonomy and alignment presents an ongoing challenge for LLM-powered multi-agent systems. Striking the right balance is crucial to ensure an efficient and effective task-execution process that faithfully accomplishes the user-defined \texttt{Goal}.

In the following Section \ref{sec:taxonomy}, we explain how our taxonomy addresses these challenges of analyzing and balancing the interplay between autonomy and alignment across architectural aspects and viewpoints.

\section{Multi-Dimensional Taxonomy}
\label{sec:taxonomy}

In this section, we introduce the system of our multi-dimensional taxonomy, engineered to methodically analyze the interplay between autonomy and alignment across architectures of autonomous LLM-powered multi-agent systems. The taxonomy weaves three crucial dimensions, i.e.\ levels of autonomy, levels of alignment as well as architectural viewpoints. Together, they form a three-dimensional matrix, serving as a comprehensive repository of architectural design options (see Fig.\ \ref{figure:cube}). 

Section \ref{subsec:tax-balancing} delves into the complexities of the interplay between autonomy and alignment, exploring how the synergies of different levels of autonomy and alignment can characterize a system's dynamics.
Subsequently, Section \ref{subsec:tax-viewpoints}, underscores the importance of incorporating architectural viewpoints into the taxonomic system. Rather than applying the autonomy-alignment matrix flatly, we propose analyzing each architectural viewpoint as well as further inherent architectural characteristics individually.
Such a viewpoint-focused approach allows for a deeper and more nuanced understanding of the systems, reflecting the complexity of their architectural design and resulting dynamics.
Finally, in Section \ref{subsec:tax-matrices}, we unify these components, mapping the autonomy-alignment dimensions and levels onto the identified architectural viewpoints, thus introducing a third dimension into our taxonomy. Furthermore, we distinguish architectural aspects inherent to these viewpoints and specify corresponding level criteria for both autonomy and alignment.

This framework provides a comprehensive classification of autonomous LLM-powered multi-agent systems, revealing distinct insights into the complexities arising from the interplay of interdependent architectural aspects. Each aspect is characterized by its levels of autonomy and alignment, influencing the systems' behavior, interactions, composition, and interaction with contextual resources.

\subsection{Interplay between Autonomy and Alignment}
\label{subsec:tax-balancing}
Autonomy and alignment, as interdependent and interplaying concepts, have their roots in management sciences and organizational behavior, playing integral roles in the ways teams and systems function \cite{mintzberg1989structuring,o2008ambidexterity}. In these fields, autonomy typically refers to the degree of discretion employees or teams possess over their tasks, while alignment denotes the degree to which these tasks correspond to the organization's overall objectives. 

When shifting focus to the AI landscape, the interplay between autonomy and alignment remains pivotal \cite{russell2015research,bostrom2017superintelligence}. AI systems, by nature, operate with varying degrees of independence and are often designed to accomplish tasks that are multifaceted, interconnected, and potentially beyond the capabilities of individual human operators. However, complete autonomy can pose risks. If the goals of an AI system deviate from those of its human supervisors, it could lead to unforeseen consequences or uncontrollable side effects. Therefore, controlling the level of autonomy is crucial to maintain the balance between operational efficiency and safety.
As such, understanding and defining the bounds of autonomy and alignment becomes vital to appropriately guide system behavior and prevent unwanted consequences, especially when dealing with autonomous multi-agent systems powered by LLMs. 

In order to efficiently address the characteristics of current and forthcoming autonomous LLM-powered multi-agent systems, we adopt a pragmatic and technical perspective on both autonomy and alignment. In the following sections, we explain this perspective and elaborate in detail on the dimensions and levels of autonomy and alignment applied by our taxonomy. Table \ref{table:alignment_autonomy} gives an overview of the employed levels and the resulting spectrum of potential combinations.

\begin{table}[ht]

\renewcommand{\arraystretch}{1.5}
\begin{small}
\sffamily
\centering
\begin{tabularx}{\columnwidth}{p{3cm}|X|X|X}
\textbf{Levels of Autonomy \& Alignment}& \cellcolor[HTML]{c2cfee}\textbf{L0: Static} & \cellcolor[HTML]{c2cfee}\textbf{L1: Adaptive} & \cellcolor[HTML]{c2cfee}\textbf{L2: Self-Organizing} \\
\hline
\cellcolor[HTML]{c2cfee}\textbf{L2: Real-time \newline Responsive} & \circled{3} User-Supervised \newline Automation & \circled{6} User-Collaborative \newline Adaptation & \circled{9} User-Responsive \newline Autonomy \\
\hline
\cellcolor[HTML]{c2cfee}\textbf{L1: User-Guided} & \circled{2} User-Guided \newline Automation & \circled{5} User-Guided \newline Adaptation & \circled{8} User-Guided \newline Autonomy \\
\hline
\cellcolor[HTML]{c2cfee}\textbf{L0: Integrated} & \circled{1} Rule-Driven Automation & \circled{4} Pre-Configured \newline Adaptation & \circled{7} Bounded Autonomy \\

\end{tabularx}
\end{small}
\caption{Matrix showcasing the interplay between gradations of alignment (\textit{vertical}) and autonomy (\textit{horizontal}) in the context of LLM-powered multi-agent architectures. Each cell signifies a unique combination of autonomy and alignment levels. Refer to Sections \ref{subsubsec:autonomy} and \ref{subsubsec:alignment} for details on the applied levels. Section \ref{subsec:tax-aa-matrix} provides details on the resulting matrix combinations.}
\label{table:alignment_autonomy}
\end{table}

\subsubsection{Autonomy}
\label{subsubsec:autonomy}
The degree of autonomy refers to the extent to which an AI system can make decisions and act independently of rules and mechanisms defined by humans. For LLM-powered multi-agent systems, this translates to a system's proficiency in addressing the goals or tasks specified by the user in a self-organizing manner, adapting and re-calibrating to the complexities of a given situation.
Autonomous multi-agent systems are \textit{by nature} striving for this end-to-end automatic goal completion and task management from a user perspective.
While automation often gets conflated with autonomy, it's essential to differentiate the two. 
Automation pertains to tasks being carried out without human input \cite{brustoloni1991autonomous,sae2016taxonomy}, while autonomy pertains to \textit{decisions about tasks} being made without human intervention \cite{franklin1996agent,parasuraman2000model,beer2014toward}. In the domain of LLM-powered multi-agent systems, we look beyond mere task automation, focusing on how these systems internally manage their dynamics to fulfill user objectives.
Our taxonomy, therefore, distinguishes systems on a spectrum of autonomy. Drawing from the triadic interplay illustrated in Fig.\ \ref{figure:triadic-relationship}, on the one end of the spectrum, we see systems that heavily rely on predefined rules and frameworks, set by their human system architects. While they may execute tasks autonomously, their decision-making process is constrained within a fixed set of parameters (\textit{low autonomy}).
On the other hand, we encounter systems characterized by their ability for self-organisation and dynamic self-adaptation. Rather than relying on hard-coded mechanisms, they harness the power of LLMs to interpret, decide, and act, making them more adaptable to changing situations (\textit{high autonomy}).

\textbf{Autonomy Levels.}
The levels of autonomy, represented on the x-axis in our matrix (see Fig.\ \ref{figure:cube} and Table \ref{table:alignment_autonomy}), articulate the degree of agency of the LLM-powered agents in making decisions regarding the system operation independently from predefined and automated mechanisms.

\begin{itemize}
    \item [\textbf{L0:}] \textbf{Static Autonomy} - At this foundational level, systems are primarily automated, relying heavily on the rules, conditions, and mechanisms embedded by system architects. The systems follow defined rules and predetermined mechanisms. This, however, includes some degree of flexibility resulting from rule-based options and alternatives. Anyway, the agents in these systems, are not empowered to modify rules during runtime. For instance, their function here is limited to the effective execution of assigned tasks. Depending on the alignment level, this results in Rule-Driven Automation, User-Guided Automation, or User-Supervised Automation (see Table \ref{table:alignment_autonomy}).

    \item [\textbf{L1:}] \textbf{Adaptive Autonomy} - Evolving from the static level, systems at this stage possess the capability to adapt their behavior within a structure and procedural guidelines established by the system architects. The LLM-powered agents are capable of adjusting the system's operations within this provided framework (such as flexible infrastructures and protocols) due to the needs of the given application scenarios, but not beyond. Depending on the alignment level, this leads to Pre-Configured Adaptation, User-Guided Adaptation, or User-Collaborative Adaptation.

    \item [\textbf{L2:}] \textbf{Self-Organizing Autonomy} - At this highest level of autonomy,  LLM-powered agents emerge as the principal actors, capable of self-organization, actively learning and dynamically tailoring their operations in real-time based on environmental cues and experiences. The autonomy lies not in being independent from user intervention, but in being independent of architect-defined rules and mechanisms. However, this might also include highly generic infrastructures that are modifiable by the LLM-powered agents and thus allow self organisation. Depending on the alignment level, this results in Bounded Autonomy, User-Guided Autonomy, or User-Responsive Autonomy.
\end{itemize}

These levels of autonomy not only apply to the system as a whole, but to architectural viewpoints and involved architectural characteristics (see Section \ref{subsec:tax-matrices}).

\subsubsection{Alignment}
\label{subsubsec:alignment}
In the context of AI, the term alignment traditionally refers to the challenge of ensuring that an AI system's behavior aligns with human intentions, values or goals. This intricate problem, often framed as the \textit{control problem}, is a cornerstone of AI safety discourse \cite{bostrom2017superintelligence,russell2019human}.
However, when viewed through a practical lens, especially in the context of autonomous LLM-powered multi-agent systems, the alignment paradigm acquires a more interactive, user-centric perspective \cite{amodei2016concrete}. 
Here, alignment techniques can be seen as a detailed calibration of conditions tied to user-specified objectives or complex tasks. This includes preferences, policies, constraints, and boundaries which collectively steer or regulate the system's trajectory towards achieving its set targets. Importantly, within this framework, alignment is not seen as counter to autonomy. Instead, it acts to complement and refine it, being applicable across various levels of autonomy.

It is also important to note that these alignment aspects are focused on the agent-interaction layer and do not, or at least only indirectly, concern the utilized large language models (LLMs) \cite{askell2021general,wolf2023fundamental} or other contextual resources, such as foundation models. 
However, the agent-interaction layer extends alignment possibilities by integrating rules and mechanisms to control agent interactions, for example, by incorporating real-time monitoring via \textit{interceptors} \cite{bass2003software}.  Such measures, as delineated by \cite{laplante2004real,hellerstein2004feedback} enable precise control over agent interactions as well as their interactions with LLMs and contextual resources, ensuring that they adhere to predetermined conditions and behaviors.
Moreover, employing methodologies like \textit{design by contract} \cite{meyer1992applying} further augments this control. Through this paradigm, software components' formal, verifiable specifications, or \textit{contracts}, can delineate conditions for agent interactions, especially when they engage with foundational resources like LLMs. Such contracts can specify acceptable behaviors, constraints, and criteria, ensuring the system behaves reliably and as intended.
While foundational models like LLMs have their inherent challenges, the agent-interaction layer introduces a distinct dimension of complexity. It is imperative to ensure both are seamlessly and securely integrated, ensuring alignment across all levels of the system.

For our taxonomy, we combine two important dimensions of alignment: its origin and timing, reflecting the dynamic tension between automated alignment mechanisms and human customization, as illustrated in Fig.\ \ref{figure:triadic-relationship}.
The origin delves into who dictates the alignment, the system architect or the system user. Meanwhile, timing refers to when the alignment is specified, encompassing phases like pre-deployment, post-deployment but prior to runtime, or even during runtime.
Autonomous LLM-powered multi-agent systems strive for achieving a goal or accomplishing a task prompted by the system user. Given this user-centric model, the alignment techniques that are integrated into the system architecture might address generic aspects, which are not directly related to the nuances and characteristics of specific user goals. 
To address this, we've categorized alignment into levels. The base level, or \textit{low alignment level}, signifies alignment that's already embedded into the system's design by the system architects. This intrinsic alignment sets broad behavioral boundaries without focusing on specific user preferences. On the other hand, the \textit{high alignment levels} are more adaptable and centered around user-specified alignment. Here, users have the flexibility to set their preferences either before the system enters its runtime or, ultimately, during its active operation. This dynamic range ensures a tighter fit to user objectives, all however built upon mechanisms integrated into the system architecture. 

\textbf{Alignment Levels.}
The levels of alignment, represented on the y-axis in our matrix (see Fig.\ \ref{figure:cube} and Table \ref{table:alignment_autonomy}), measure the degree to which users of the system can influence or adjust the system's behavior.

\begin{itemize}
    \item [\textbf{L0:}] \textbf{Integrated Alignment} - At this foundational level, the alignment techniques are built directly into the system's architecture. In such system, alignment mechanisms are static and rule-driven, and cannot be altered by the users. Depending on the autonomy level, this results in Rule-Driven Automation, Pre-Configured Adaptation, or Bounded Autonomy, where user interaction with the system's alignment is not provided (see Table \ref{table:alignment_autonomy}).
    
    \item [\textbf{L1:}] \textbf{User-Guided Alignment} - Evolving from the previous level, the User-Guided Alignment offers a degree of customization. This level empowers users by allowing them to set or adjust specific alignment parameters, such as conditions, rules, or boundaries, before the system starts its operation. These interactions are primarily facilitated via user interfaces designed to capture user preferences in a structured manner. Depending on the autonomy level, this results in User-Guided Automation, User-Guided Adaptation, or User-Guided Autonomy.
    
    \item [\textbf{L2:}] \textbf{Real-Time Responsive Alignment} - The highest level of alignment is represented by means to adjust the system's behavior in real-time. Thanks to integrated real-time monitoring mechanisms, the system can actively solicit user feedback user decisions at critical junctures or decision points. This responsiveness enables a high level of collaboration in terms of ongoing feedback between the user and the system. Depending on the autonomy level, this results in User-Supervised Automation, User-Collaborative Adaptation, or User-Responsive Autonomy.
\end{itemize}

The hierarchical alignment structure mirrors the challenges commonly faced in software development, especially as visualized by the \textit{cone of uncertainty} \cite{boehm1988understanding}. This cone depicts how uncertainties, predominant in the early stages of a project, gradually diminish as developers gain better clarity.
Transferred to the alignment of LLM-powered multi-agent systems, initial alignment challenges are approached with a broad brushstroke by system architects. Their main focus is to ensure foundational system functionality. At this juncture, the specificity of user-driven tasks, with their unique nuances and intricacies, will hardly be anticipated by system architects.
In turn, a user-guided, pre-runtime alignment allows users to specify preferences and limitations based on a more concrete understanding of possible challenges associated to a given task. However, even at this stage, it is barely possible for the user to anticipate all alignment challenges.
Factors like ambiguous prompts, incomplete task specifications, or, in general, unclear expectations, can inadvertently steer the system off course resulting in unintended outcomes and uncontrollable side effects.
Thus, once the system is in operation, such deviations from user's intentions might first become obvious to the user during runtime and require adjustment in real-time. This allows the system to re-align based on immediate user feedback. 

These levels of alignment not only apply to the system as a whole, but to architectural viewpoints and involved architectural characteristics (see Section \ref{subsec:tax-matrices}).

\subsubsection{Combinations of Autonomy and Alignment}
\label{subsec:tax-aa-matrix}
\begin{figure}[htb]
	\centering
	\includegraphics[width= 0.48\textwidth]{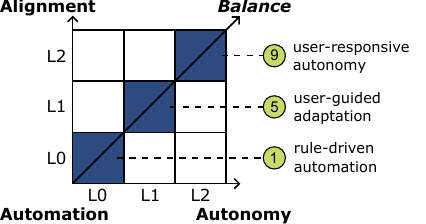}
	\caption{Interplay between autonomy and alignment: balancing evolving levels of dynamism and responsibilities of both LLM-powered agents (\textit{autonomy}) and human users (\textit{alignment}).}
	\label{figure:balanced-matrix}
\end{figure}

By combining these two dimensions in our matrix (see Table \ref{table:alignment_autonomy}), we provide a comprehensive view of the interplay between diverse gradations of autonomy and alignment within LLM-powered multi-agent systems.
As illustrated in Fig.\ \ref{figure:balanced-matrix}, departing from static and rule-driven system configurations (\textit{automation}), this autonomy-alignment matrix captures the progression of dynamism and responsibilities as we move along the axes. 
On the y-axis, alignment levels represent the gradation of human users' involvement—from integrated systems where the user's role is passive (L0), to real-time responsive setups demanding active participation (L2). On the y-axis, the autonomy levels signify the evolving capabilities of LLM-powered agents, progressing from static behaviors (L0) to adaptive (L1) and, ultimately, self-organizing mechanisms (L2). 
This matrix structure reflects the triadic interplay and dynamic tensions illustrated in Fig.\ \ref{figure:triadic-relationship}. 
As we delve deeper into the matrix, the challenge becomes evident: ensuring balance between the evolving responsibilities of LLM-powered agents and the goals and intentions by the human users, ultimately resulting in a dynamic collaboration between agents and humans. 

In the following, we detail the resulting nine combinations provided in Table \ref{table:alignment_autonomy}.

\begin{enumerate}

    \item [\circled{1}] \textbf{Rule-Driven Automation (L0 Autonomy/L0 Alignment)}: In this configuration, both autonomy and alignment are at the lowest levels. Such system operates based on scripted mechanisms and fixed conditions defined by the system architects. The alignment aspects are integrated into the system during the development stage. At this level, the behavior can not be affected by the user, neither pre-runtime or during runtime. However, this \textit{balanced} setup is ideal for repetitive, well-defined tasks that require minimal variability or adaptability.

    \item [\circled{2}] \textbf{User-Guided Automation (L0 Autonomy/L1 Alignment)}: Here, while the autonomy of the system remains at the lowest level, users can guide the system's behavior within predefined parameters before runtime. The user can not make real-time adjustments, but can influence the system's behavior within the predefined structure. It allows a certain level of customization without granting complete control, which can be ideal for scenarios where user expertise can refine the operation but the task management remains static.

    \item [\circled{3}] \textbf{User-Supervised Automation (L0 Autonomy/L2 Alignment)}: This configuration allows the user to supervise and make real-time adjustments to the system, despite the system's autonomy level remaining at the lowest. The user has more control over the system's behavior, being able to guide and correct it as necessary in real-time. This configuration is suitable for tasks requiring real-time user feedback and supervision, but where the processes themselves are performed in a pre-scripted manner.

    \item [\circled{4}] \textbf{Pre-Configured Adaptation (L1 Autonomy/L0 Alignment)}: At this level combination, the multi-agent system can adapt its behavior within certain predefined parameters, but the alignment aspects are still integrated into the system during the development stage, with no room for adjustments by the user during runtime. This allows the agents to handle a greater variety of scenarios than strictly rule-driven systems, while still maintaining a clear boundary on its behavior set by the predefined parameters.

    \item [\circled{5}] \textbf{User-Guided Adaptation (L1 Autonomy/L1 Alignment)}: Here, the system can adapt its operations within predefined parameters, and the user can also guide the system's behavior within a predefined structure. It offers a \textit{balanced} mix of system adaptation and user guidance. This combination can be useful when the tasks or environment have some level of unpredictability that requires the LLM-powered agents to adapt within predefined bounds, and where the user’s guidance can inform the system's decisions.

    \item [\circled{6}] \textbf{User-Collaborative Adaptation (L1 Autonomy/L2 Alignment)}: This configuration allows the system to adapt its operations and also be responsive to real-time user adjustments. It offers a dynamic interaction between the user and the agents. This configuration is well-suited to environments that are unpredictable and require the system to adapt and respond quickly to the user's real-time instructions.

    \item [\circled{7}] \textbf{Bounded Autonomy (L2 Autonomy/L0 Alignment)}: Here, the multi-agent system can self-organize and learn from the environment, but the alignment is integrated during the development stage and cannot be adjusted by users during runtime. This provides the system with a great degree of flexibility to handle complex tasks and environments, while still adhering to a defined set of limitations and constraints specified by the system architect.

    \item [\circled{8}] \textbf{User-Guided Autonomy (L2 Autonomy/L1 Alignment)}: At this level, while the system can self-organize and learn from the environment, the user can guide the system's behavior within predefined parameters. This system configuration leverages the agents' self-organizing abilities while allowing user guidance pre-runtime. It combines the strengths of autonomous decision making and learning, with the assurance of user-specified boundaries.

    \item [\circled{9}] \textbf{User-Responsive Autonomy (L2 Autonomy/L2 Alignment)}: This is the highest level of autonomy and alignment, where the LLM-powered agents can self-organize, learn from the environment and user's real-time adjustments. It offers a \textit{balanced} collaborative environment between the user and the agents, being ideal for complex, unpredictable environments where both autonomous strategy and action as well as real-time user-responsiveness are needed.

\end{enumerate}

In the following sections, we explore how the autonomy-alignment matrix can be applied within the context of architectural viewpoints and further architectural aspects inherent to these viewpoints on autonomous LLM-powered multi-agent systems.

\subsection{Architectural Viewpoints} 
\label{subsec:tax-viewpoints}

Architectural viewpoints are a structured means to analyze and assess complex systems from diverse perspectives focusing on selected aspects and layers of an architecture \cite{bass2003software,clements2003documenting}. Central to these viewpoints is the consideration of stakeholder concerns, which inform and determine the highlighted aspects and their interrelations in each viewpoint. Providing a combined multi-perspective analysis, viewpoints serve as an effective framework to examine the structures and dynamics of software architectures. For our taxonomy, we leverage viewpoints on autonomous LLM-powered multi-agent systems. Rather than mapping the autonomy-alignment taxonomy flatly onto the system, which oversimplifies the multi-faceted nature of these systems, analyzing each architectural viewpoint individually offers a tailored lens, enabling to comprehend the role and impact of autonomy and alignment within the system. Each viewpoint reveals distinct insights into the system's behavior, internal interactions, composition, and context interaction, leading to a more nuanced and comprehensive classification \cite{rozanski2012software}. 
\begin{figure}[htb]
	\centering
	\includegraphics[width= 0.45 \textwidth]{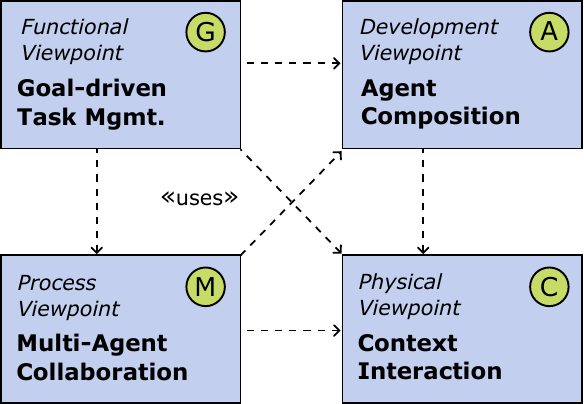}
	\caption{Architectural viewpoints oriented to the \textit{4+1 view model of software architecture} \cite{kruchten19954+} applied to autonomous LLM-powered multi-agent systems.
 }
	\label{figure:viewpoints}
\end{figure}

\subsubsection{Applied Viewpoints}
\label{subsubsec:viewpoints}
For our taxonomy, we orient to Kruchten's renowned \textit{4+1 view model of software architecture} \cite{kruchten19954+}, an established standard viewpoint model for software architecture, adapting it to suit the architectural characteristics of LLM-powered multi-agent systems (see Section \ref{sec:architecture}). Our taxonomy encompasses the following four architectural viewpoints on these systems (refer to Fig.\ \ref{figure:viewpoints}):

\begin{itemize}
    \item [\circled{G}] \textbf{Goal-driven Task Management} \textit{(Functional Viewpoint)}: Kruchten's functional viewpoint refers to the system's visible functionalities as experienced by its users \cite{kruchten19954+}. In the context of autonomous LLM-powered multi-agent systems, we see \texttt{Goal-driven Task Management} as a manifestation of this functional viewpoint. It entails the system's capabilities and mechanisms to decompose user-prompted goals or complex tasks into smaller, more manageable tasks, and subsequently, orchestrate task execution, combine the results, and deliver the final result forming the response for the user (see Figs.\ \ref{figure:architecture} and \ref{figure:ontology}).
    
    \item [\circled{A}] \textbf{Agent Composition} \textit{(Development Viewpoint)}: According to Kruchten, the development viewpoint is primarily focusing on the system's software architecture, the breakdown into components, and their organization \cite{kruchten19954+}. In our context, we interpret this as \texttt{Agent Composition}, focusing on the system's internal composition, particularly the assembly and constellation of agents. It includes the types and roles of agents, their memory usage, the relationships between agents (see Figs.\ \ref{figure:architecture} and \ref{figure:ontology}).
    
    \item [\circled{M}] \textbf{Multi-Agent Collaboration} \textit{(Process Viewpoint)}: Kruchten's process viewpoint concerns the dynamic aspects of a system, specifically the system procedures and interactions between components \cite{kruchten19954+}. We apply this to the \texttt{Multi-Agent Collaboration} in our model, emphasizing the collaborative task execution and interactions among agents. This encompasses the application of communication protocols, the dynamics of actions management, such as the actual task execution, mutual task delegation, as well as the evaluation and merging of task results on agent level, as well as the management of communication components such as prompts and prompt templates (see Figs.\ \ref{figure:architecture} and \ref{figure:ontology}).
    
    \item [\circled{C}] \textbf{Context Interaction} \textit{(Physical Viewpoint)}: According to Kruchten, the physical viewpoint involves the system's mapping to physical resources \cite{kruchten19954+}. We extend this to \texttt{Context Interaction}, focusing on the system's interaction with the external environment. It includes how the system acquires, integrates, and utilizes contextual resources such as external data, expert tools, and further foundation models as well as the organized distribution and utilization of contextual resources within the agent network (see Figs.\ \ref{figure:architecture} and \ref{figure:ontology}). 
\end{itemize}

\subsubsection{Viewpoint Interdependencies}
\label{subsubsec:dependencies}
To effectively design and understand autonomous LLM-powered multi-agent systems, it's essential to recognize the relationships and interdependencies between architectural components and viewpoints \cite{zhao2001using, rozanski2012software}.
Fig.\ \ref{figure:viewpoints} illustrates these interrelated architectural viewpoints for multi-agent architectures. The figure includes \textit{use} dependencies between the viewpoints, denoted as dotted lines indicating the directions of usage \cite{UML}. 
These dependencies arise from the interconnected nature of these viewpoints, as they collectively shape the behavioral features provided by the system, here expressed in terms of the \texttt{Goal-driven Task Management} viewpoint. 

\begin{figure}[htb]
	\centering
	\includegraphics[width= \textwidth]{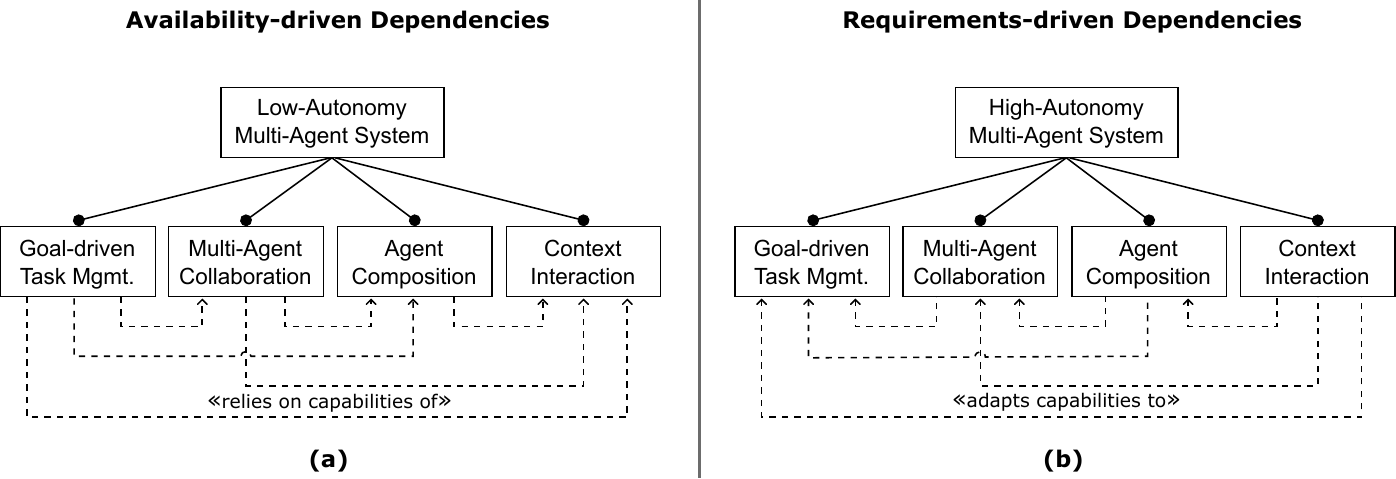}
	\caption{Types of dependencies distinguished by different levels of autonomy provided by LLM-powered multi-agent architectures.}
	\label{figure:dependency-types}
\end{figure}

The interplay between architectural viewpoints is notably influenced by the autonomy levels of the systems. With regard to the levels of autonomy, we can distinguish the following two types of dependencies between architectural viewpoints, i.e., \textit{availability-driven dependencies} for low-autonomy systems and \textit{requirements-driven dependencies} for high-autonomy systems. Fig.\ \ref{figure:dependency-types} illustrates the two types in a simplified manner.
For further details on dependencies between architectural aspects inherent to the viewpoints, also see the feature diagram in Fig.\ \ref{figure:feature-diagram}.

\textbf{Availability-driven Dependencies (Low-Autonomy System).}
For low-autonomy multi-agent systems, as depicted in Fig.\ \ref{figure:dependency-types} (a), the architecture operates predominantly under pre-established automation. In these systems, functionality largely relies on pre-configured rules and mechanisms. Thus, the functionality of such multi-agent system is contingent upon the predefined capabilities of the system processes, which are defined by the structure of the system and the resources available. 

\begin{itemize}
    \item In such systems, \texttt{Goal-driven Task Management} depends on all other dimensions, as it represents the culmination of the system's operations in terms of the key functionality as perceived by the user. The capabilities regarding the decomposition and orchestration of task execution towards completing the prompted task are essentially influenced by the predefined composition and constellation of agents (such as competencies, roles, types, and network), their scripted mode of collaboration for task execution, and the rule-based integration and utilization of contextual resources.
    \item In turn, \texttt{Multi-Agent Collaboration} derives its operational modus from the foundational structures established by the \texttt{Agent Composition} and \texttt{Context Interaction}. The mode of collaboration for task execution among agents is dictated by predefined characteristics and competencies (types, roles) of the agents involved and their relationships and organization as network (\texttt{Agent Composition}), as well as by the accessibility in terms of the effective integration and utilization of contextual information, tools and models to execute the given tasks (\texttt{Context Interaction}).
    \item Finally, \texttt{Agent Composition} also relies on the availability of contextual resources (Context Interaction). The types and number of agents needed in the system as well as their roles and competencies are directly influenced by the contextual environment (e.g., the availability, accessibility, and quality of data, foundation models, and expert tools) used by the system and how they are utilized within the task-management activity.
\end{itemize}

\textbf{Requirements-driven Dependencies (High-Autonomy System).}
In turn, high-autonomy multi-agent systems, illustrated in Fig.\ \ref{figure:dependency-types} (b), have the ability to self-organize. In these systems, the architectural infrastructure and dynamics as well as the context interaction are self-organizing and thus capable of adapting their capabilities to the needs and requirements set by a given goal. Thus, compared to systems with low autonomy, there is an inverse dependency relationship.

\begin{itemize}
    \item In highly-autonomous multi-agent systems, the user-prompted goal delineates the requirements, charting the course for the entire architectural edifice of the system. 
    All other viewpoints adapt to the envisioned functional behavior expressed as \texttt{Goal-driven Task Management}. Based on the complexity of the goal, its decomposition into tasks and their distribution, the other architectural aspects inherent to the three further viewpoints undergo adaptations to fit the needs of the given situation.
    \item In addition, \texttt{Agent Composition}, encompassing agents' roles, types, and their memory usage, as well as \texttt{Context Interaction}, including the integration and utilization of resources, adapt to the requirements set by the modes of collaboration to tackle the assigned tasks, including the communication protocol (\texttt{Multi-Agent Collaboration}). 
    \item Finally, also \texttt{Agent Composition} sets the requirement for the adaptation of \texttt{Context Interaction}. The specific roles of agents within the system mandate particular resource integration. For instance, an agent with analytical responsibilities might necessitate the inclusion of specific data streams or computational tools. 
\end{itemize}

\textbf{Intertwined Dependencies (Mixed Autonomy Levels).}
The two distinguished types, availability-driven and requirements-driven dependencies, address the challenge of interconnected architectural viewpoints in an illustrative, but simplified manner.
On the one hand, the viewpoints of a multi-agent system might provide different autonomy levels; on the other hand, also the aspects or mechanisms within a viewpoint might be on different levels of autonomy. Both cases result in \textit{intertwined dependencies}. 
It is important to note that the autonomy levels set for one viewpoint or aspects can have an impact on others viewpoints or aspects due to their interconnected nature (see above).
The intertwined dependencies might prove risky. They can introduce complexities and unpredictabilities, potentially jeopardizing system efficiency and effectiveness. It underscores the necessity of incorporating robust control mechanisms to navigate and manage these interdependencies, which is illustrated by the following illustrative example. 

\textbf{Example.} Consider a practical scenario where an autonomous LLM-powered multi-agent system is operating in the following dynamic environment:
\begin{itemize}
    \item \textbf{Decomposition Dynamics}: Within the \texttt{Goal-driven Task Management} viewpoint, tasks are dynamically decomposed into sub-tasks based on user requirements, and this decomposition operates with a L2 autonomy level, which signifies a self-organizing manner.
    \item \textbf{Agent Collaboration Dynamics}: Similarly, the \texttt{Multi-Agent Collaboration} viewpoint, which encompasses how agents collaborate for task execution, operates on the same L2 autonomy. Agents decide on-the-fly how to interact, delegate tasks, and merge results.
    \item \textbf{Contextual Interaction Limitation}: Contrasting the above, the \texttt{Context Interaction} viewpoint is constrained by L0 autonomy level. Here, the system's access and interaction with the external environment (contextual resources) are limited by predefined rules. The system cannot autonomously decide to reach out to new resources or modify the way it interacts with existing ones.
    
\end{itemize}

Given these conditions, a potential issue arises: As tasks are decomposed and agents plan their collaborations, they might, based on their L2 autonomy, decide to utilize certain contextual resources. However, when it's time to access these resources, they might find them inaccessible due to the L0 constraints in the \texttt{Context Interaction} viewpoint. This discrepancy in autonomy levels can cause operational dead-ends. For instance, an agent might anticipate using an external data source to complete its task, but the L0 constraints prevent it from accessing that source, leaving the task incomplete. 

Such issues highlight the importance of having robust control mechanisms in place that can preemptively identify and mitigate these discrepancies, ensuring smooth system operations.

For a detailed illustration of dependencies between viewpoint-specific aspects, refer to Section \ref{subsubsec:ex-vp-aspects}.

\begin{table*}[ht]
\renewcommand{\arraystretch}{1.5}
\begin{scriptsize}
\sffamily
\centering
\begin{tabularx}{\textwidth}{p{0.7cm}|X|X|X|X|X}
\textbf{Matrix \#} & \textbf{Autonomy-Alignment \newline Levels}& \cellcolor[HTML]{c2cfee}\circled{G} \textbf{Goal-driven \newline Task Management} & \cellcolor[HTML]{c2cfee}\circled{A} \textbf{Agent \newline Composition} & \cellcolor[HTML]{c2cfee}\circled{M} \textbf{Multi-Agent \newline Collaboration} & \cellcolor[HTML]{c2cfee}\circled{C} \textbf{Context \newline Interaction} \\ 
\hline
\cellcolor[HTML]{c2cfee}\circled{1}&\cellcolor[HTML]{c2cfee}\textbf{Rule-Driven Automation:} Static \& Integrated (L0 \& L0) & Rule-driven task management. &  Rule-driven agent composition and constellation. & Rule-driven collaboration protocols. & Rule-driven interaction with contextual resources. \\ 
\hline
\cellcolor[HTML]{c2cfee}\circled{2}&\cellcolor[HTML]{c2cfee}\textbf{User-Guided Automation:} Static \& User-Guided (L0 \& L1) & User-guided task management. & User-guided agent composition and constellation. & User-guided collaboration protocols. & User-guided context integration and utilization. \\ 
\hline
\cellcolor[HTML]{c2cfee}\circled{3}&\cellcolor[HTML]{c2cfee}\textbf{User-Supervised Automation:} Static \& Real-Time Responsive (L0 \& L2) & Task management adjusted during runtime. & Agent composition and constellation adjusted during runtime. & Agent collaboration adjusted during runtime. & Context integration and utilization adjusted during runtime. \\ 
\hline
\cellcolor[HTML]{c2cfee}\circled{4}&\cellcolor[HTML]{c2cfee}\textbf{Pre-Configured Adaptation:} Adaptive \& Integrated (L1 \& L0) & Adaptive task management with predefined options. & Adaptive agent composition and constellation with predefined flexibility. & Adaptive collaboration protocols. & Pre-integrated adaptive contextual resources. \\ 
\hline
\cellcolor[HTML]{c2cfee}\circled{5}&\cellcolor[HTML]{c2cfee}\textbf{User-Guided Adaptation:} Adaptive \& User-Guided (L1 \& L1) & User-adjusted adaptive task management. & User-adjusted adaptive agent composition and constellation. & User-adjusted adaptive collaboration. & User-adjusted adaptive context integration and utilization. \\ 
\hline
\cellcolor[HTML]{c2cfee}\circled{6}&\cellcolor[HTML]{c2cfee}\textbf{User-Collaborative Adaptation:} Adaptive \& Real-Time Responsive (L1 \& L2) & Adaptive task management adjusted during runtime. & Adaptive agent composition and constellation adjusted during runtime. & Adaptive collaboration adjusted during runtime. & Adaptive context integration and utilization adjusted during runtime. \\ 
\hline
\cellcolor[HTML]{c2cfee}\circled{7}&\cellcolor[HTML]{c2cfee}\textbf{Bounded Autonomy:} Self-Organizing \& Integrated (L2 \& L0) & Task management organically based on current needs. & Agents self-organize based on current scenario. & Collaboration strategy evolves organically. & Agents select from a pool of contextual resources. \\ 
\hline
\cellcolor[HTML]{c2cfee}\circled{8}&\cellcolor[HTML]{c2cfee}\textbf{User-Guided Autonomy:} Self-Organizing \& User-Guided (L2 \& L1) & User-guided self-organizing task management. & User-guided agent self-organization. & User-guided collaboration evolution. & User-guided self-organized selection from contextual resources. \\ 
\hline
\cellcolor[HTML]{c2cfee}\circled{9}&\cellcolor[HTML]{c2cfee}\textbf{User-Responsive Autonomy:} Self-Organizing \& Real-Time Responsive (L2 \& L2) & Self-organizing task management adjusted during runtime. & Agent self-organization adjusted during runtime. & Collaboration evolution adjusted during runtime. & Self-organized selection from contextual resources adjusted during runtime. \\ 
\end{tabularx}
\end{scriptsize}
\caption{Mapping autonomy and alignment levels (\textit{vertical}, \#1--9 resulting from Table \ref{table:alignment_autonomy}) to architectural viewpoints (\textit{horizontal}) on autonomous LLM-powered multi-agent systems resulting in 36 viewpoint-specific system configurations. A detailed explanation of the autonomy and alignment levels is provided in Section \ref{subsec:tax-balancing}. For an overview of the applied viewpoints, refer to Section \ref{subsec:tax-viewpoints}.}
\label{table:taxonomy-application}
\end{table*}

\subsection{Interplay of Autonomy and Alignment in the System Architecture} 
\label{subsec:tax-matrices}

As already illustrated, both autonomy and alignment serve as \textit{cross-cutting concerns} \cite{kiczales1997aspect} impacting the operational efficiency of various architectural aspects across LLM-powered multi-agent systems. Thus, in the following, we map our matrix of autonomy and alignment levels onto the architectural viewpoints. This projection crafts a three-dimensional matrix, offering a prism through which these systems can be analyzed and categorized (also see Fig.\ \ref{figure:cube}). In Section \ref{subsubsec:map-to-viewpoints}, we give systematic overview of the resulting viewpoint-specific combinations of autonomy and alignment levels. Section \ref{subsubsec:ex-vp-aspects} details the architectural aspects associated to these viewpoints and specifies corresponding level criteria that establish the foundation for the taxonomic classification.

\subsubsection{Mapping Autonomy-Alignment Levels to Viewpoints}
\label{subsubsec:map-to-viewpoints}

Table \ref{table:taxonomy-application} showcases the interplay of autonomy, alignment, and the distinct architectural viewpoints. It applies the autonomy-alignment matrix, as illustrated in Table \ref{table:alignment_autonomy}, to the identified architectural viewpoints inherent to autonomous LLM-powered multi-agent systems. Each cell in this matrix signifies a unique architectural design choice, representing a distinct system configuration.
The architectural viewpoints (\textit{horizontal}; see Section \ref{subsec:tax-viewpoints}) are categorized into \texttt{Goal-driven Task Management}, which highlights the system's functionalities; \texttt{Agent Composition}, emphasizing its intrinsic structure; \texttt{Multi-Agent Collaboration}, denoting the dynamics of agent interactions; and \texttt{Context Interaction}, detailing the system's rapport with its external environment in terms of data and tools.
Alongside these viewpoints, the nine combinations of autonomy and alignment levels (\textit{vertical}; see Section \ref{subsec:tax-balancing}) describe the system's behavior. Autonomy ranges from \texttt{Static} to \texttt{Self-Organizing}, determining the system's degree of self-organization. Meanwhile, alignment varies from \texttt{Integrated} to \texttt{Real-Time Responsive}, capturing the depth of human influence over the system's operations. Combining these dimensions results in 36 system architectural design options available for configuring multi-agent systems.

In the following Section \ref{subsubsec:ex-vp-aspects}, we explore further viewpoint-specific aspects and their interdependencies, in order to derive level criteria for the taxonomic classification.

\subsubsection{Viewpoint-specific Aspects and Level Criteria}
\label{subsubsec:ex-vp-aspects}

As outlined above, architectural viewpoints provide means to analyze certain aspects and aspect relations of the system's architecture in a multi-perspective manner \cite{rozanski2012software}. Drawing from the domain-ontology model (Fig.\ \ref{figure:ontology}), we now systematize the viewpoint-specific aspects employed in our taxonomy. Subsequently, we specify level criteria for autonomy and alignment corresponding to each aspect. Furthermore, we outline the main interdependencies among these aspects.  

\begin{figure}[htb]
	\centering
	\includegraphics[width= \textwidth]{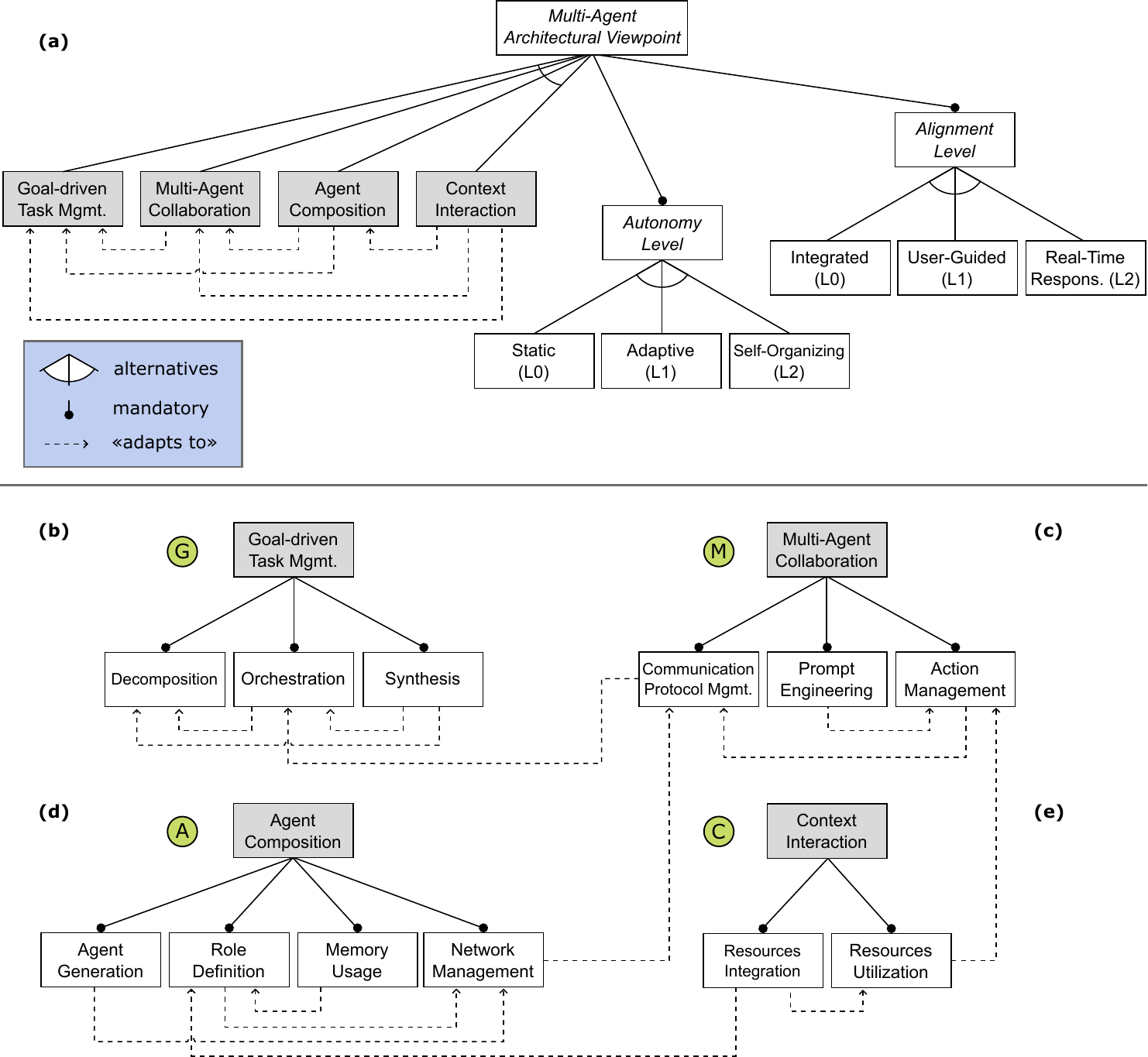}
    \caption{Feature diagram showcasing the taxonomic structure. Each viewpoint integrates autonomy and alignment levels \textbf{(a)}. The diagram further illustrates viewpoint-specific aspects and mechanisms \textbf{(b--e)} alongside the \textit{adapts-to} dependencies among them.
    }
	\label{figure:feature-diagram}
\end{figure}

Fig.\ \ref{figure:feature-diagram} gives an overview of our taxonomy's characteristics, structured through a feature diagram \cite{batory2005feature,schobbens2007generic}. 
Employed predominantly in software engineering, feature diagrams visually express feature models, 
which aim to organize the hierarchical structure as well as dependencies among system features.

In particular, Fig.\ \ref{figure:feature-diagram} (a) structures the viewpoint-specific taxonomic structure. Each of the four integrated viewpoints provides a certain combination of autonomy and alignment levels. As illustrated in Figs.\ \ref{figure:feature-diagram} (b--e), this structure is refined by viewpoint-specific aspects and their interdependencies in terms of requirements-driven dependencies (\textit{adapts-to}), presuming a high-autonomy system configuration, as discussed in Section \ref{subsubsec:dependencies}. These dependencies suggest that the capabilities of a dependent aspect evolve in line with the needs and stipulations of the aspect it points to. In turn, also these viewpoint-specific aspects can be assessed by the autonomy and alignment levels, resulting in a more nuanced taxonomic classification.

Across the four distinct viewpoints, a total of 12 characteristic aspects are identified (as illustrated in Fig.\ \ref{figure:feature-diagram}). Each of these aspects can be assessed and classified by its corresponding autonomy and alignment levels, yielding 9 possible configuration options per aspect (detailed in Table \ref{table:alignment_autonomy}). Thus, given the viewpoints \(V_1, V_2, V_3, V_4\) with respective aspect counts \(A_1 = A_2 = 3\), \(A_3 = 4\), and \(A_4 = 2\), and a level count \(L = 3\) for both autonomy and alignment, we define:

\begin{align}
T_A &= \sum_{i=1}^{4} A_i, &\text{(Total Aspects)} \\
S_C &= L^2, &\text{(Single Configuration Options per Aspect)} \\
T_{SC} &= T_A  S_C, &\text{(Total Single Configuration Options)} \\
T_{CC} = (L^2)^{A_1 + A_2 + A_3 + A_4} &= S_C^{T_A}. &\text{(Total Combined Configurations)}
\end{align}

Using the provided values, we find \(T_{SC} = 108\) and \(T_{CC} = 9^{12} \approx 282 \times 10^9\).

In sum, mapping the autonomy-alignment matrix onto the identified aspects, our taxonomy captures 108 distinct single configuration options. When considering all possible combinations of these configurations, we arrive at a total of \( 9^{12} \), which equates to roughly 282 billion combinations available for configuring LLM-powered multi-agent architectures. This underscores the complexity challenge posed by such systems, further accentuated by the various options for intertwined dependencies, as detailed in Section \ref{subsubsec:dependencies}.

In the following, we outline these viewpoint-specific aspects, drawing from the architectural specifications detailed in Section \ref{subsec:aa-ontology} and define corresponding criteria for the levels of autonomy and alignment.

\circled{G} \textbf{Aspects and Levels of Goal-driven Task Management.} 

Taxonomic aspects of \texttt{Goal-driven Task Management} comprise the three constituting phases: \texttt{Decomposition} (how the goal or complex task is broken down into manageable sub-tasks), \texttt{Orchestration} (how these tasks are distributed among the LLM-powered agents), and \texttt{Synthesis} (how the results of the tasks are finally combined); refer to Fig.\ \ref{figure:feature-diagram} (b). 

\textbf{Level Criteria:}
    \begin{itemize}
        \item \textbf{Static Autonomy (L0)}: At this level, we observe scripted processes and automated mechanics with rule-based options and alternatives for the task-management activity, including the phases of task decomposition, distributing and orchestrating the execution of single tasks, or combining their results. These scripted automated processes might demonstrate variability and flexibility including iterations based on predetermined mechanics and conditions. However, this level also includes strict processes or execution chains with no variations.
        
        \item \textbf{Adaptive Autonomy (L1)}: Here, the system provides predefined but adaptive procedures for the phases of the task-management activity. Based on these predefinitions integrated into the system's design and implementation, the LLM-powered agents are vested with certain autonomy to adapt the managing and controlling of the task-management processes. For example, within a defined framework, the agents are involved in managing the task decomposition, the distribution to other agents, or decisions about the synthesis of results. For this purpose, also patterns or prepared mechanics are reused.
        
        \item \textbf{Self-Organizing Autonomy (L2)}: This level embodies the LLM-powered agents' capability to architect and implement their own strategy for deconstructing and solving problems due to the characteristics or complexity of a given goal. This might also include high-level generic frameworks scaffolding the agents' interactions and processes, but leaving space to LLM-powered agents for effectively self-organizing the phases of the task-management process. 
        
        \item \textbf{Alignment Levels}: At this juncture, alignment can be seen in terms of information and constraints regarding the task-management activity, especially regarding the mechanics in the decomposition, orchestration or synthesis of sub-processes, e.g., decomposition depth or consensus options for the total result. The alignment is either integrated into the system's design (L0), configurable by the user before runtime (L1) or adjustable during runtime (L2).
    \end{itemize}

\circled{M} \textbf{Aspects and Levels of Multi-Agent Collaboration.}

For the taxonomic classification within \texttt{Multi-Agent Collaboration}, we consider \texttt{Communication-Protocol Management} (how the collaboration and dialogues between the agents are managed), \texttt{Prompt Engineering} (how prompts are applied during collaboration and executing the actions), and \texttt{Action Management} (how the different kinds of action, such as the delegation or actual execution of tasks, or the result evaluation, performed by the agents are managed); see Fig.\ \ref{figure:feature-diagram} (c)).

\textbf{Level Criteria:}    
    \begin{itemize}
        \item \textbf{Static Autonomy (L0)}: At this level, collaborative actions and interactions among agents adhere to a fixed script or set of rules. The communication protocols, prompt use and augmentation, as well as the management of actions are pre-defined and don't adjust dynamically based on agent inputs or environmental changes. Agents communicate, delegate tasks, execute instructions, and evaluate results based strictly on established, non-adaptable guidelines. Variability in the collaboration process is minimal and doesn't account for unforeseen scenarios or complexities.

        \item \textbf{Adaptive Autonomy (L1)}: This level introduces adaptability of collaboration aspects for LLM-powered agents based on predefined mechanisms. For example, the communication protocol, the prompt templates, or the management of the agent actions are modifiable. While the foundational mechanisms are preset, the LLM-powered agents can autonomously select and adapt them due to the evolving requirements of the given scenario. For this purpose, they might reuse prepared mechanisms or patterns. 

        \item \textbf{Self-Organizing Autonomy (L2)}: Agents operating at this level showcase the capability to independently strategize their collaboration for task execution. Driven by the specific demands of the set goals and task complexities, these LLM-powered agents actively plan and execute collaboration strategies that best address the scenario at hand. For instance, LLM-powered agents can self-organize protocols for collaboration, mechanisms of prompt engineering and negotiate collaboratively the execution of actions among the agent network.
        
        \item \textbf{Alignment Levels:} Relevant considerations include information and constraints linked to collaboration mechanisms and patterns between agents, the specification of prompt templates, constraints for prompt augmentation, or preferences for the execution of actions. These components can be either embedded within the system's design (L0), made available for user configuration before runtime (L1), or be amenable to real-time adjustments (L2).
    \end{itemize}

\circled{A} \textbf{Aspects and Levels of Agent Composition.}

The aspects of \texttt{Agent Composition} applied by the taxonomy comprise \texttt{Agent Generation} (how the agents are created, including the strategies and mechanisms employed), \texttt{Role Definition} (how agents' roles are specified), \texttt{Memory Usage} (how the agents utilize their memory, i.e., how information is summarized and stored, or how memory is used for reflecting instructions or planning actions), and \texttt{Network Management} (how the constellation and relationships among agents are managed); refer to Fig.\ \ref{figure:feature-diagram} (d).

\textbf{Level Criteria:} 
    \begin{itemize}
        \item \textbf{Static Autonomy (L0)}: This level features a predefined and rule-driven composition and constellation of agents. Rules and mechanisms manage the creating of agents, select the agent types, and delineate their roles and competencies. Memory utilization follows predefined mechanisms, as well as the relationship between agents.
        
        \item \textbf{Adaptive Autonomy (L1)}: While a system at this level provides predefined structures, it grants a degree of flexibility, permitting LLM-powered agents to adapt their composition and constellation within the given framework and due to given scenarios. For example, agents can replicate instances, their competencies are extensible and roles and further attributes (such as the size or compression mode for the agent memory) can be modified. Agents can modify or extend existing relationships, e.g., by connecting with further agents.
        \item \textbf{Self-Organizing Autonomy (L2)}: LLM-powered agents operating at this level exhibit the ability to autonomously define and generate types and establish collaborative networks. The impetus for such self-organization arises from an acute understanding of the demands and nuances of the given scenario. Instead of adhering to predefined agent types and roles or relationships, agents dynamically constitute and organize based on real-time needs. 
        \item \textbf{Alignment Levels:} Pertinent to agent composition are information and constraints regarding their creation, types, roles, and competencies. Further, the manner in which agents interrelate and how they are structured within the network holds significance. These mechanics and configurations can be either deeply embedded into the system's design (L0), be made configurable by the user before runtime (L1), or be dynamically adjustable during system operation (L2).
    \end{itemize}

\circled{C} \textbf{Aspects and Levels of Context Interaction.}

For \texttt{Context Interaction}, the taxonomic aspects comprise (\texttt{Resources Integration} (how the integration of contextual resources in terms of data, tools, models, and other applications is achieved), and \texttt{Resources Utilization} (how these resources are actually utilized for executing tasks); refer to Fig.\ \ref{figure:feature-diagram} (e).

\textbf{Level Criteria:} 
    \begin{itemize}
        \item \textbf{Static Autonomy (L0)}: At this level, contextual resources, including data, expert tools, and specialized foundation models, are rigidly integrated based on the system's initial design. Their utilization is organized by predefined rules and patterns relating scenarios and resource application. However, this level also includes the case that certain or any resources might not be available for use.
        \item \textbf{Adaptive Autonomy (L1)}: Certain contextual resources are pre-integrated, but the system provides adaptive mechanisms usable by LLM-powered agents for integrating missing resources when needed. To this end, access to certain APIs might be prepared. Based on predetermined mechanisms, the LLM-powered agents can flexible determine how to best utilize and combine these provided resources, tailoring their approach to the unique requirements of the given scenario.
        \item \textbf{Self-Organizing Autonomy (L2)}: LLM-powered agents possess the autonomy to interface with a diverse pool of contextual resources (cf. \textsc{Hugging Face}). They can discerningly select, integrate, and harness these resources based on the objectives at hand and the specific challenges they encounter.
        \item \textbf{Alignment Levels:} Factors to consider encompass information and constraints pertaining to the integration and application of contextual resources. These may include specifications or guidelines on which resources to leverage, when and how to integrate them, any limitations on their utilization, and more. These specifications or guidelines might be built into the system's design (L0), made available for user modification prior to runtime (L1), or even be adapted in real time (L2).
    \end{itemize}

In the following Section \ref{sec:comparison}, we explore the application of our taxonomy to real-world LLM-based multi-agent systems. 

\section{Classification of Selected Systems}
\label{sec:comparison}

In order to demonstrate the practical utility of our taxonomy, we analyze and classify selected existing autonomous LLM-powered multi-agent systems. We have chosen a set of seven state-of-the-art multi-agent systems for this assessment: \textsc{AutoGPT} \cite{autogpt2023}, \textsc{BabyAGI} \cite{babyagi2023}, \textsc{SuperAGI} \cite{superagi2023}, \textsc{HuggingGPT} \cite{shen2023hugginggpt}, \textsc{MetaGPT} \cite{hong2023metagpt}, \textsc{CAMEL} \cite{li2023camel}, and \textsc{AgentGPT} \cite{agentgpt2023}. Each of these systems is maintained and available as open-source project. For basic information on these and further LLM-powered multi-agent systems, refer to Section \ref{subsec:bg-architectures}. 
For each selected system, we gathered relevant information by examining the technical documentation and research papers, where available, as well as reviewing the code base. We further engaged with each system to explore its real-time functionalities, with emphasis on alignment mechanisms available before and during runtime.

In the following sections, we first report on the results of analyzing and classifying the selected systems (Section \ref{subsec:comp-classification}). Then, we compare and interpret the results in Section \ref{subsec:comp-interpretation}.

\begin{table}[h]
\renewcommand{\arraystretch}{2}
\begin{scriptsize}
\sffamily
\centering
\setlength\tabcolsep{2.65pt} 
\begin{tabular}{l||*{24}{c|}}
\cline{2-25}
\multirow{3}{*}{
\parbox[c]{2cm}{\centering\textbf{LLM-powered Multi-Agent Systems}}} & \multicolumn{6}{c|}{\cellcolor[HTML]{c2cfee}\textbf{Goal-driven Task Mgmt.}} & \multicolumn{6}{c|}{\cellcolor[HTML]{c2cfee}\textbf{Multi-Agent Collaboration}} & \multicolumn{8}{c|}{\cellcolor[HTML]{c2cfee}\textbf{Agent Composition}} & \multicolumn{4}{c|}{\cellcolor[HTML]{c2cfee}\textbf{Context Interact.}} \\
\cline{2-25}
& \multicolumn{2}{c|}{\cellcolor[HTML]{c2cfee}\textbf{Decom}} & \multicolumn{2}{c|}{\cellcolor[HTML]{c2cfee}\textbf{Orch}} & \multicolumn{2}{c|}{\cellcolor[HTML]{c2cfee}\textbf{Synth}} & \multicolumn{2}{c|}{\cellcolor[HTML]{c2cfee}\textbf{CommP}} & \multicolumn{2}{c|}{\cellcolor[HTML]{c2cfee}\textbf{PrEng}} & \multicolumn{2}{c|}{\cellcolor[HTML]{c2cfee}\textbf{ActM}} & \multicolumn{2}{c|}{\cellcolor[HTML]{c2cfee}\textbf{AGen}} & \multicolumn{2}{c|}{\cellcolor[HTML]{c2cfee}\textbf{RoleD}} & \multicolumn{2}{c|}{\cellcolor[HTML]{c2cfee}\textbf{MemU}} & \multicolumn{2}{c|}{\cellcolor[HTML]{c2cfee}\textbf{NetM}} & \multicolumn{2}{c|}{\cellcolor[HTML]{c2cfee}\textbf{Integ}} & \multicolumn{2}{c|}{\cellcolor[HTML]{c2cfee}\textbf{Util}} \\
\cline{2-25}
& \cellcolor[HTML]{c2cfee}\textbf{AU} & \cellcolor[HTML]{c2cfee}\textbf{AL} & \cellcolor[HTML]{c2cfee}\textbf{AU} & \cellcolor[HTML]{c2cfee}\textbf{AL} & \cellcolor[HTML]{c2cfee}\textbf{AU} & \cellcolor[HTML]{c2cfee}\textbf{AL} & \cellcolor[HTML]{c2cfee}\textbf{AU} & \cellcolor[HTML]{c2cfee}\textbf{AL} & \cellcolor[HTML]{c2cfee}\textbf{AU} & \cellcolor[HTML]{c2cfee}\textbf{AL} & \cellcolor[HTML]{c2cfee}\textbf{AU} & \cellcolor[HTML]{c2cfee}\textbf{AL} & \cellcolor[HTML]{c2cfee}\textbf{AU} & \cellcolor[HTML]{c2cfee}\textbf{AL} & \cellcolor[HTML]{c2cfee}\textbf{AU} & \cellcolor[HTML]{c2cfee}\textbf{AL} & \cellcolor[HTML]{c2cfee}\textbf{AU} & \cellcolor[HTML]{c2cfee}\textbf{AL} & \cellcolor[HTML]{c2cfee}\textbf{AU} & \cellcolor[HTML]{c2cfee}\textbf{AL} & \cellcolor[HTML]{c2cfee}\textbf{AU} & \cellcolor[HTML]{c2cfee}\textbf{AL} & \cellcolor[HTML]{c2cfee}\textbf{AU} & \cellcolor[HTML]{c2cfee}\textbf{AL}\\
\hline \hline
\cellcolor[HTML]{c2cfee}\textbf{Auto-GPT} \cite{autogpt2023} & \cellcolor[HTML]{E5E5E5}2&0& \cellcolor[HTML]{E5E5E5}0&0& \cellcolor[HTML]{E5E5E5}1&0& \cellcolor[HTML]{E5E5E5}0&0& \cellcolor[HTML]{E5E5E5}1&0& \cellcolor[HTML]{E5E5E5}2&0& \cellcolor[HTML]{E5E5E5}0&0& \cellcolor[HTML]{E5E5E5}1&0& \cellcolor[HTML]{E5E5E5}0&0& \cellcolor[HTML]{E5E5E5}0&0& \cellcolor[HTML]{E5E5E5}0&0& \cellcolor[HTML]{E5E5E5}2&0 \\
\hline
\cellcolor[HTML]{c2cfee}\textbf{BabyAGI} \cite{babyagi2023} & \cellcolor[HTML]{E5E5E5}2&0& \cellcolor[HTML]{E5E5E5}0&0& \cellcolor[HTML]{E5E5E5}1&0& \cellcolor[HTML]{E5E5E5}0&0& \cellcolor[HTML]{E5E5E5}1&0& \cellcolor[HTML]{E5E5E5}2&0& \cellcolor[HTML]{E5E5E5}0&0& \cellcolor[HTML]{E5E5E5}1&0& \cellcolor[HTML]{E5E5E5}0&0& \cellcolor[HTML]{E5E5E5}0&0& \cellcolor[HTML]{E5E5E5}0&0& \cellcolor[HTML]{E5E5E5}2&0 \\
\hline
\cellcolor[HTML]{c2cfee}\textbf{SuperAGI} \cite{superagi2023} & \cellcolor[HTML]{E5E5E5}2&0& \cellcolor[HTML]{E5E5E5}1&0& \cellcolor[HTML]{E5E5E5}1&1& \cellcolor[HTML]{E5E5E5}0&0& \cellcolor[HTML]{E5E5E5}1&0& \cellcolor[HTML]{E5E5E5}2&0& \cellcolor[HTML]{E5E5E5}1&1& \cellcolor[HTML]{E5E5E5}2&1& \cellcolor[HTML]{E5E5E5}0&1& \cellcolor[HTML]{E5E5E5}0&0& \cellcolor[HTML]{E5E5E5}0&1& \cellcolor[HTML]{E5E5E5}2&1 \\
\hline
\cellcolor[HTML]{c2cfee}\textbf{HuggingGPT} \cite{shen2023hugginggpt} & \cellcolor[HTML]{E5E5E5}2&0& \cellcolor[HTML]{E5E5E5}1&0& \cellcolor[HTML]{E5E5E5}2&0& \cellcolor[HTML]{E5E5E5}0&0& \cellcolor[HTML]{E5E5E5}2&0& \cellcolor[HTML]{E5E5E5}2&0& \cellcolor[HTML]{E5E5E5}2&0& \cellcolor[HTML]{E5E5E5}2&0& \cellcolor[HTML]{E5E5E5}1&0& \cellcolor[HTML]{E5E5E5}0&0& \cellcolor[HTML]{E5E5E5}2&0& \cellcolor[HTML]{E5E5E5}2&0 \\
\hline
\cellcolor[HTML]{c2cfee}\textbf{MetaGPT} \cite{hong2023metagpt} & \cellcolor[HTML]{E5E5E5}2&0& \cellcolor[HTML]{E5E5E5}0&0& \cellcolor[HTML]{E5E5E5}2&0& \cellcolor[HTML]{E5E5E5}1&0& \cellcolor[HTML]{E5E5E5}1&0& \cellcolor[HTML]{E5E5E5}2&0& \cellcolor[HTML]{E5E5E5}0&0& \cellcolor[HTML]{E5E5E5}0&0& \cellcolor[HTML]{E5E5E5}0&0& \cellcolor[HTML]{E5E5E5}1&0& \cellcolor[HTML]{E5E5E5}0&0& \cellcolor[HTML]{E5E5E5}2&0 \\
\hline
\cellcolor[HTML]{c2cfee}\textbf{CAMEL} \cite{li2023camel} & \cellcolor[HTML]{E5E5E5}2&0& \cellcolor[HTML]{E5E5E5}0&0& \cellcolor[HTML]{E5E5E5}1&0& \cellcolor[HTML]{E5E5E5}0&0& \cellcolor[HTML]{E5E5E5}1&0& \cellcolor[HTML]{E5E5E5}1&0& \cellcolor[HTML]{E5E5E5}0&1& \cellcolor[HTML]{E5E5E5}1&1& \cellcolor[HTML]{E5E5E5}0&0& \cellcolor[HTML]{E5E5E5}0&1& \cellcolor[HTML]{E5E5E5}0&0& \cellcolor[HTML]{E5E5E5}0&0 \\
\hline
\cellcolor[HTML]{c2cfee}\textbf{AgentGPT} \cite{agentgpt2023} & \cellcolor[HTML]{E5E5E5}2&1& \cellcolor[HTML]{E5E5E5}1&0& \cellcolor[HTML]{E5E5E5}1&0& \cellcolor[HTML]{E5E5E5}0&0& \cellcolor[HTML]{E5E5E5}1&0& \cellcolor[HTML]{E5E5E5}2&0& \cellcolor[HTML]{E5E5E5}1&1& \cellcolor[HTML]{E5E5E5}2&0& \cellcolor[HTML]{E5E5E5}0&0& \cellcolor[HTML]{E5E5E5}0&0& \cellcolor[HTML]{E5E5E5}0&0& \cellcolor[HTML]{E5E5E5}2&1 \\
\hline
\cellcolor[HTML]{c2cfee}\textbf{Zapier*} \cite{rahmati2017ifttt} & \cellcolor[HTML]{E5E5E5}1&1& \cellcolor[HTML]{E5E5E5}0&1& \cellcolor[HTML]{E5E5E5}0&1& \cellcolor[HTML]{E5E5E5}0&0& \cellcolor[HTML]{E5E5E5}0&1& \cellcolor[HTML]{E5E5E5}0&1& \cellcolor[HTML]{E5E5E5}0&0& \cellcolor[HTML]{E5E5E5}0&0& \cellcolor[HTML]{E5E5E5}0&0& \cellcolor[HTML]{E5E5E5}0&0& \cellcolor[HTML]{E5E5E5}0&1& \cellcolor[HTML]{E5E5E5}0&1 \\

\hline
\end{tabular}
\end{scriptsize}
\vspace{5pt} 
\caption{Assessment of autonomy (\texttt{AU}) and alignment (\texttt{AL}) levels across viewpoint-specific aspects of selected LLM-powered multi-agent systems. Detailed level criteria for viewpoint-specific aspects are discussed in Section \ref{subsubsec:ex-vp-aspects}. * \textsc{Zapier}, a workflow-automation tool, has been included to contrast the results.}
\label{table:llm_aspect_classification}
\end{table}

\subsection{Taxonomic Classification}
\label{subsec:comp-classification}

The taxonomic classification relies on a detailed assessment of autonomy and alignment levels for viewpoint-specific aspects of the systems. Table \ref{table:llm_aspect_classification} reports on the results of assessing these levels of autonomy (\texttt{AU}) and alignment (\texttt{AL}) for aspects characterizing the four architectural viewpoints applied by our taxonomy. In particular, for \texttt{Goal-driven Task Management}, the aspects of decomposition (\texttt{Decom}), orchestration (\texttt{Orch}), and synthesis (\texttt{Synth}); for \texttt{Multi-Agent Collaboration}, the aspects of communication-protocol management (\texttt{CommP}), prompt engineering (\texttt{PrEng}), and action management (\texttt{ActM}); for \texttt{Agent Composition}, the aspects of agent generation (\texttt{AGen}), role definition (\texttt{RoleD}), memory usage (\texttt{MemU}), and network management (\texttt{NetM}); for \texttt{Context Interaction}, the aspects of resource integration (\texttt{Integ}), and resource utilization \texttt{Util} are distinguished. An overview of these viewpoint-specific aspects and corresponding level criteria applied for this assessment is provided in Section \ref{subsubsec:ex-vp-aspects}. 

* In order to contrast the classification results, we included \textsc{Zapier} \cite{rahmati2017ifttt} into our comparison, a renowned tool offering the automation of workflows based on user-specified tasks.

\begin{figure*}[htb]
	\centering
	\includegraphics[width= \textwidth]{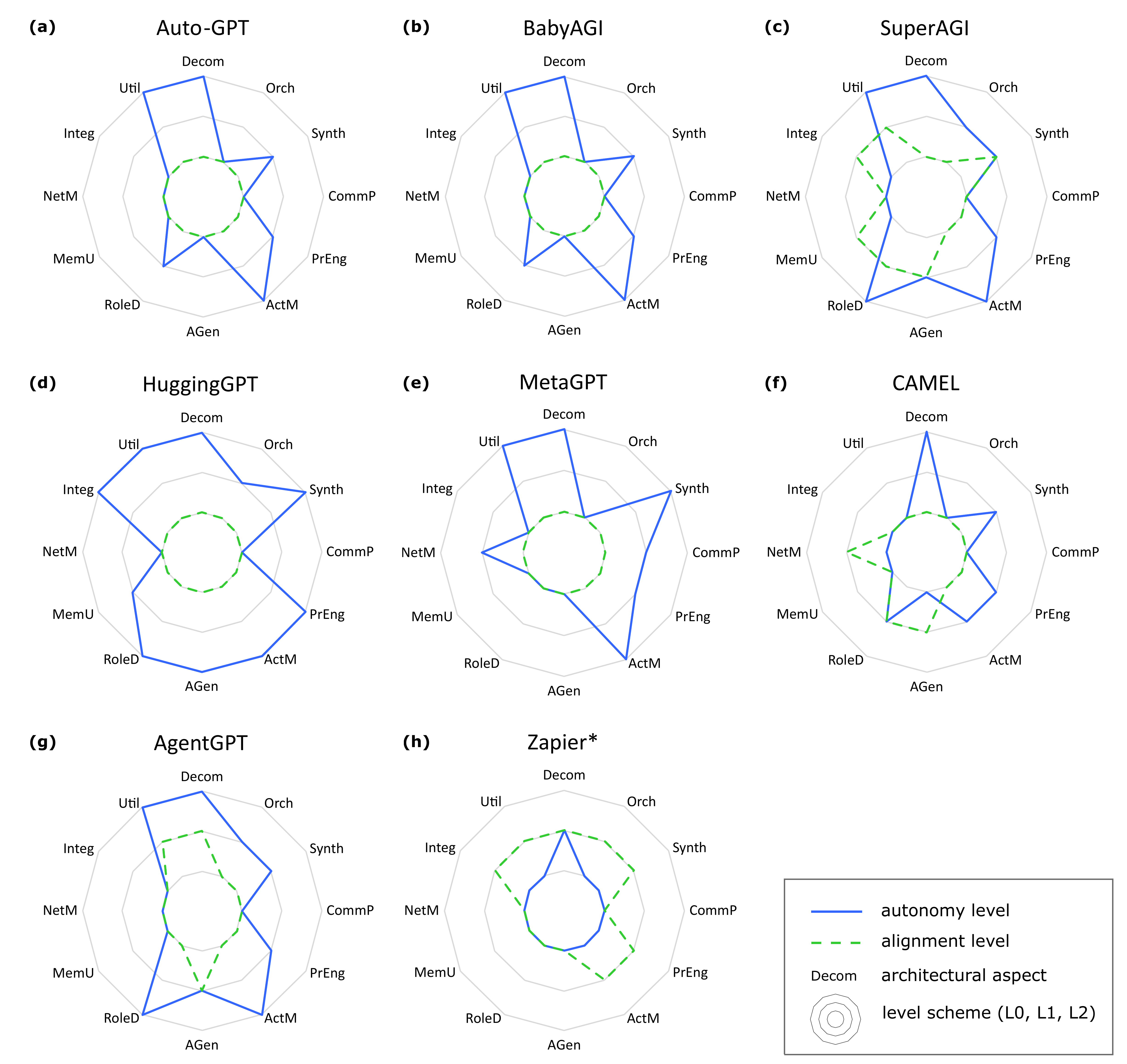}
	\caption{Radar charts illustrating the system profiles based on an assessment of architectural aspects in terms of autonomy (\textit{blue graph}) and alignment (\textit{green dashed graph}) levels. Detailed assessment data can be found in Table \ref{table:llm_aspect_classification}.}
	\label{figure:radar-aspects}
\end{figure*}

Fig.\ \ref{figure:radar-aspects} displays the derived autonomy and alignment levels per multi-agent system using radar (or spider) charts \cite{tufte2001visual}. In particular, architectural aspects form the multiple axes. The level scheme (\texttt{L0}, \texttt{L1}, \texttt{L2}) for autonomy and alignment is depicted by grey circles linking these axes. The blue graph then represents the assessed autonomy levels, the green dashed graph the corresponding alignment levels. 

In what follows, we outline key results from the taxonomic assessment for each system.

\begin{itemize}
    \item \textbf{\textsc{Auto-GPT}} \cite{autogpt2023} - enables users to specify multiple goals, which are autonomously decomposed into tasks and then prioritized (L2 autonomy for \texttt{Decomposition}); also see Fig.\ \ref{figure:radar-aspects} (a). The system encompasses three distinct task-management agents: an execution agent, a task-creation agent, and a task-prioritization agent. All tasks are actually performed by this singular execution agent sequentially determined by prioritization (L0 autonomy for \texttt{Orchestration}). Following the completion of each task, the agent evaluates the intermediate results, engaging in self-criticism. The tasks are optionally re-prioritized. The final result represents an aggregate of all partial results, complemented with a succinct summary (L1 autonomy \texttt{Synthesis}). The communication between the three agents follows a predefined communication protocol. \texttt{Prompt Engineering} is adaptive based on templates (L1 autonomy). The management of the performed actions is self-organized (L2 autonomy). Agents in the system are pre-configured and instantiated once, showing L0 autonomy. The \texttt{Role} of the execution agent is adaptive (L1 autonomy). Both \texttt{Memory Usage} and \texttt{Network Management} adhere to predefined rules, marking L0 autonomy. \textsc{Auto-GPT} is equipped with a suite of predefined contextual resources (L0 autonomy), which are utilized in a self-organizing manner due to the needs of the scenario, demonstrating L2 autonomy. Across the aspects, the system provides a low level of user interaction. Beyond the transmission of goals, further user interaction is only available in terms of authorizing the subsequent execution step, which however, can also be skipped via the \textit{continuous mode}, leaving users with no further intervention capabilities.

    \item \textbf{\textsc{BabyAGI}} \cite{babyagi2023} - provides very similar functionalities and architectural characteristics to those demonstrated by \textsc{Auto-GPT}. This similarity can be visually observed in the radar charts illustrated in Figs.\ \ref{figure:radar-aspects} (a) and (b). In particular, both systems maintain a high-autonomy level regarding goal \texttt{Decomposition}, the management of single actions performed by the task-execution agent, and the \texttt{Utilization} of integrated contextual resources due to the requirements of the given tasks. Furthermore, both systems provide transparency by reporting on the execution agent's plans and thinking operations as basis for executing the tasks. They both maintain iterative, but fixed communication protocols allowing for the task-management agents to organize the decomposition and aspects of result synthesis. \textsc{BabyAGI} extends the set of task-management agents by a context agent, which is responsible for context-interaction tasks. In addition, it also allows the user to configure a few constraints governing the use of the LLM.

    \item \textbf{\textsc{SuperAGI}} \cite{superagi2023} - also allows users to specify a goals or complex task, which are decomposed autonomously into tasks tackled sequentially by an LLM-powered agents. Thereby, its functionalities and architectural characteristics are in some regards similar to \textsc{Auto-GPT} and \textsc{BabyAGI} (see above). For instance, the executing agent is performing the assigned tasks autonomously, showcasing L2 autonomy in \texttt{Action Management}; also see Fig.\ \ref{figure:radar-aspects} (c). The systems leverages predefined, but adaptable prompt templates (L1 autonomy for \texttt{Prompt Engineering}). Contextual resources are also utilized in an autonomous manner, indicative of L2 autonomy. 
    Diverging from its counterparts, \textsc{SuperAGI} necessitates that users create a dedicated agent for every distinct goal. In contrast to the other two systems, the roles of these agents are highly task-adaptive (L2 autonomy) and can be influenced by user, attributing to L1 alignment. Moreover, the \texttt{Orchestration} of tasks is more adaptive (L1 autonomy). 
    A distinctive feature of \textsc{SuperAGI} is its ability to incorporate various alignment strategies. For instance, it permits constraints regarding \texttt{Memory Usage}, allowing users to cap the context window's length, a trait of L1 alignment. In terms of \texttt{Context Interaction}, \textsc{SuperAGI} comes with an array of pre-configured tools, all of which can be authorized for \texttt{Utilization}. Furthermore, data can be uploaded, empowering agents to seamlessly incorporate and utilize it (L1 alignment). Though \textsc{SuperAGI} can handle multiple goals and agents, these agents work in parallel, separately. There is no actual collaboration between the agents, and no further configuration options for the user, resulting in L0 autonomy and alignment for \texttt{Communication Protocol} and \texttt{Network Management}.

    \item \textbf{\textsc{HuggingGPT}} \cite{shen2023hugginggpt} - follows a different strategy by leveraging the LLM as an autonomous controller that combines various multi-modal AI models to solve complex tasks. In this, it integrates with the \textsc{Hugging Face} platform that provides a large pool of foundation models available for utilization. This singular central LLM-powered agent, tailored to solve the given goal, is autonomous in breaking down the goal or complex task into manageable tasks (L2 autonomy for \texttt{Decomposition}) as well as in selecting, combining, and applying the appropriate models via prompting, achieving L2 autonomy for \texttt{Integration} and \texttt{Utilization} of contextual resources as well as for \texttt{Prompt Engineering}, \texttt{Action Management}, \texttt{Agent Generation}, \texttt{Role Definition}. However, not every aspect of \textsc{HuggingGPT} exhibits such high autonomy. Some procedural aspects are predefined, but adaptive to the given task, such the high-level process framework consisting of the predefined phases of planning, model selection, task execution, and response generation. Despite its autonomy in many aspects, \textsc{HuggingGPT} does not grant users any further degree of user customization, resulting in L0 alignment for all aspects; see Fig.\ \ref{figure:radar-aspects} (d). In sum, based on our autonomy-alignment matrix (see Table \ref{table:alignment_autonomy}), the system shows tendency towards \texttt{Bounded Autonomy} (\#7; L2 autonomy and L0 alignment).

    \item \textbf{\textsc{MetaGPT}} \cite{hong2023metagpt} - aims to solve complex programming tasks (specifically in Python) by leveraging the synergies of multiple collaborating LLM-powered role agents. Thereby, the framework simulates human workflows and responsibilities inherent to software-development project.
    For this purpose, the task-management activity comprises distinct phases similar to the waterfall process (such as RE, design, coding, testing), each with dedicated role agents responsible for autonomously executing the associated tasks. Each phase delivers certain artefacts then processed by the next phase (e.g., design specification). In particular, the user-specified requirements are autonomously transferred into these different artefacts (L2 autonomy for \texttt{Decomposition}), which are finally also combined to product a tested software program, achieving L2 autonomy for \texttt{Synthesis}; also see Fig.\ \ref{figure:radar-aspects} (e). As mentioned aove, the actual \texttt{Orchestration} follows a defined scheme, termed as \textit{standardized operation process}, resulting in L0 autonomy. Both \texttt{Agent Generation} and \texttt{Roles} are predefined (L0 autonomy). Within their designated phases, the agents display pronounced autonomy, exhibiting adaptability in their \texttt{Action Management} corresponding to the specificity of tasks, thereby reaching L2 autonomy. \texttt{Prompt Engineering} is predefined, but adapted for inter-agent collaboration (L1 autonomy). Contextual resources are autonomously utilized as needed, marking L2 autonomy for \texttt{Context Interaction}. Similar to \textsc{HuggingGPT}, \textsc{MetaGPT} showcases low levels of alignment for the user (L0), since it provides no further configuration or adjustment options for the user.

    \item \textbf{\textsc{CAMEL}} \cite{li2023camel} - aims to explore the potentials of autonomous cooperation among communicative LLM-powered agents to accomplish complex tasks. Similar the most other multi-agent systems, it aspires to handle given user-prompted goals autonomously. To this end, a dedicated generic task-specifier agent breaks down the goal into a list of manageable tasks (L2 autonomy for \texttt{Decomposition}), also see Fig.\ \ref{figure:radar-aspects} (f). Subsequently, these tasks are processed by a pair of agents working in tandem through a cyclical dialogue pattern, wherein the AI-user agent lays out the directives, and the AI-assistant agent assumes the role of the executor. This strict modus operandi corresponds to L0 autonomy in \texttt{Orchestration}, \texttt{Communication Protocol}, and \texttt{Network Management}. The specific \texttt{Roles} of these predefined agent archetypes can be selected by the user (L1 alignment). Augmenting this duo are other specialized agents designed for specific roles, including task allocation and strategic planning. In contrast to most other analyzed systems (except \textsc{MetaGPT}), \textsc{CAMEL} provides actual collaboration between role agents executing the given tasks. During the task-execution phases, agents operate with a marked sense of autonomy, achieving L2 in both \texttt{Prompt Engineering} and \texttt{Action Management}. Alignment options for the user are provided via the definition of agents, encompassing their \texttt{Roles} and interrelation in the \texttt{Network}, resulting in L1 alignment.

    \item \textbf{\textsc{AgentGPT}} \cite{agentgpt2023} - also strives to accomplish a user-prompted goal by leveraging a single task-execution agent, who can be created by the user by specifying its goal, resulting in L1 alignment for \texttt{Agent Generation}.  The agent systematically addresses tasks, prioritizing them based on a predefined list and capitalizing on contextual resources, all in a self-organized manner corresponding to L2 autonomy.     
    In terms of autonomy levels across different facets, it closely mirrors \textsc{SuperAGI}, as can be observed in Fig.\ \ref{figure:radar-aspects} (c) and (g). However, when it comes to alignment possibilities, \textsc{AgentGPT} diverges slightly. On the one hand, it provides no adjustment options regarding the agent's role or the \texttt{Synthesis} of results (both L0 alignment). On the other hand, it introduces the option to extend the task list by inserting custom tasks, achieving L1 alignment for \texttt{Decomposition}.

    \item \textbf{\textsc{Zapier}} \cite{rahmati2017ifttt} - Unlike the other entities discussed, \textsc{Zapier} focuses on workflow automation based on user-specified tasks and does not represent an LLM-powered multi-agent system. Its inclusion in this classification serves to contrast the results, providing a clearer understanding of the capabilities and potential limits of LLM-powered systems when juxtaposed with traditional task-oriented automation platforms. In particular, in \textsc{Zapier}, users need to define step-by-step instructions to facilitate the automation process. The resulting workflows can work in parallel, but lack the capability for direct inter-task interactions.
    \textsc{Zapier} offers configuration options (pre-runtime) for diverse aspects related to \texttt{Task Management} and \texttt{Context Interaction}, resulting L1 alignment; also see Fig.\ \ref{figure:radar-aspects} (h). Given its non-reliance on LLM-powered agents, it naturally secures an L0 ranking in both autonomy and alignment for agent-centric attributes. Nevertheless, it leverages LLMs to process textual tasks such as writing emails, or for decomposing user-specified goals into tasks, a feature users can optionally activate (thus L1 autonomy for \texttt{Decomposition}). Drawing from our autonomy-alignment matrix, detailed in Table \ref{table:alignment_autonomy}, \textsc{Zapier} is aptly categorized as \texttt{User-Guided Automation} (\#2; signifying L0 autonomy and L1 alignment).
    Given its unique positioning as a workflow automation system, \textsc{Zapier} provides an illustrative deviation from these trends. Its strategic approach is distinct, predominantly showcasing lower levels of autonomy, as the LLMs are only leveraged for specific, limited tasks. Conversely, it favors a strategy of extensive user-guided alignment, applicable to all except the agent-specific aspects. Note, however, that alignment here is not applied as an enhancement or refinement to actually align the system's operation to the user's goal or intention, but in terms of specifications of process steps with detailed instructions essential for the system's operation.

\end{itemize}

\subsection{Comparative Analysis}
\label{subsec:comp-interpretation}

In the following, we discuss the distribution of assessed levels (Section \ref{subsubsec:comparison}) and explore strategies across system categories (Section \ref{subsubsec:grouping}).

\subsubsection{Comparison of Assessed Levels}
\label{subsubsec:comparison}

\begin{figure*}[htb]
	\centering
	\includegraphics[width= 0.8 \textwidth]{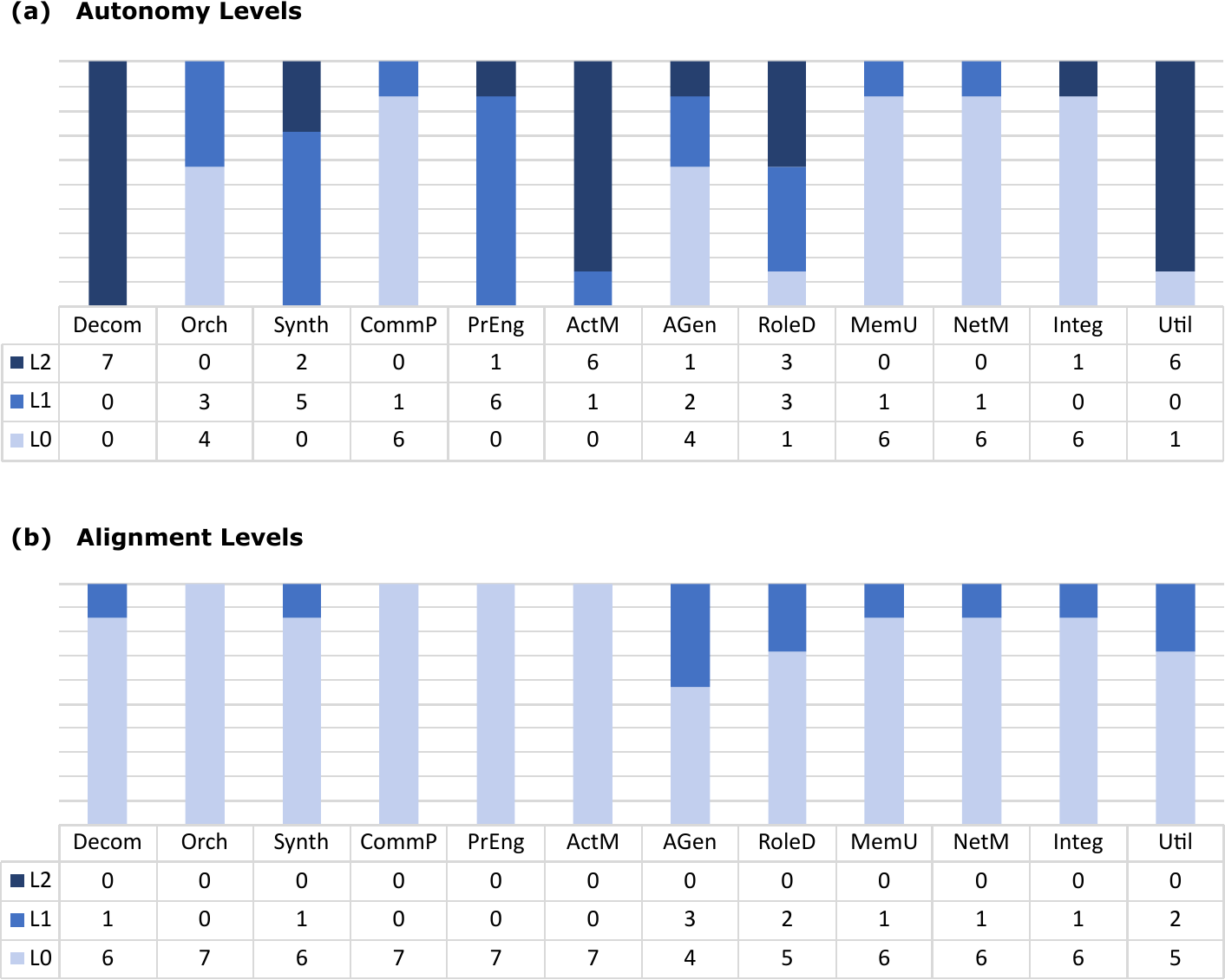}
    \caption{Distribution of identified autonomy and alignment levels across architectural aspects of selected LLM-powered multi-agent systems, represented as stacked bar charts with corresponding data provided below.}
    \label{figure:barcharts}
\end{figure*}

Fig.\ \ref{figure:barcharts} offers an overview of how the assessed levels of autonomy and alignment distribute over the 12 categories of architectural aspects of the seven selected multi-agent systems. Detailed assessment data is provided in Table \ref{table:llm_aspect_classification}.

On the one hand, three groups of aspect categories emerge when assessing autonomy levels, each displaying a certain degree of homogeneity. A detailed representation can be found in Fig.\ \ref{figure:barcharts} (a).

\begin{itemize}
    \item \textbf{High-Autonomy Aspects:} Among the systems, we encounter a high-autonomy strategy for certain aspects, demonstrated by self-organizing and autonomously deciding LLM-powered agents. This strategy is particularly evident for the decomposition of goals into manageable tasks (\texttt{Decom}), for the management of actions, encompassing the actual performance of different task-related actions (\texttt{ActM}), as well as for utilizing the contextual resources such as tools and data (\texttt{Util}). Nearly all systems delegate the responsibilities for these aspects to the LLM-powered agents, which corresponds to \texttt{L2} autonomy.

    \item \textbf{Medium-Autonomy Aspects:} For other aspects, systems lean towards a semi-autonomous strategy (\texttt{L1}), featuring predefined mechanisms adaptable by the LLM-powered agents. This is prominently observed in two aspects. First, in result synthesis (\texttt{Synth}), by combining the task results guided by a predefined framework adaptable by the LLM-powered agents. Second, in the engineering of prompts (\texttt{PrEng}), such during prompt augmentation by adapting predefined prompt templates.
    
    \item \textbf{Low-Autonomy Aspects:} Several architectural aspects showcase a deterministic strategy with rule-based mechanisms and automation, demonstrating \texttt{L0} autonomy, which can observed for the following aspects:
\begin{itemize}
    \item orchestrating and distributing the tasks (\texttt{Orch}).
    \item guiding the collaboration between the agents (\texttt{CommP}).
    \item managing the utilization of memory, such as for reflecting and planning (\texttt{MemU}).
    \item managing the agent network, such as regarding the relationships between the agents (\texttt{NetM}).
    \item integrating contextual resources (\texttt{Integ}).
\end{itemize}

\end{itemize}

\textbf{Variable-Autonomy Aspects:}
The autonomy levels for the aspects of agent generation (\texttt{AGen}) and role definition (\texttt{RoleD}) display notable variability, as depicted in Fig.\ \ref{figure:barcharts} (a). This heterogeneity is reflective of the different strategies employed by the multi-agent systems under analysis (detailed in Section \ref{subsubsec:grouping}).

\textbf{Integrated and User-Guided Alignment:} 
Drawing insights from Fig.\ \ref{figure:barcharts} (b), it emerges that the predominant strategy across most systems is to maintain lower levels of alignment across all assessed aspects. This primarily manifests in alignment techniques already integrated into the system architecture (L0 alignment), offering little to no options for user adjustment. Furthermore, low-autonomy aspects with predefined and automated mechanisms can be used to control and align other higher-autonomy levels. Thus, these mechanisms can be seen as manifestations of integrated alignment.
However, we observe a noticeable inclination for systems to provide user-guided alignment (L1) for specific aspect categories, namely the agent generation (\texttt{AGen}), agent role definition (\texttt{RoleD}), and contextual resource utilization (\texttt{Util}). Furthermore, the data reveals a consistent lack of real-time responsive alignment options across all examined systems. Nonetheless, in this context, it is worth mentioning that some systems at least facilitate monitoring functionalities available for system users (often termed as \textit{verbose mode}), which provide transparency of the reasoning and decision-making performed by the execution agents during task reflection and planning. This transparency grants users the leverage to either greenlight or halt the impending actions. However, there are no possibilities to further influence the task planning or execution, such as by adjusting or refining task planning.

\textbf{Intertwined Dependencies:}
As evident from the radar charts in Fig.\ \ref{figure:radar-aspects}, a diverse range of autonomy levels manifests both within and across architectural viewpoints of the analyzed systems. This variance results in a complex web of \textit{intertwined dependencies} between the aspects: 
Certain aspects have to deal with diverse dependencies. While dependent on predefined mechanisms or resources provided by low-autonomy aspects (\textit{availability-driven dependencies}), they have to adapt dynamically in response to situational imperatives set by other high-autonomy aspects (\textit{requirements-driven dependencies}).
This complexity resulting from intertwined dependencies can be seen as challenging for ensuring accurate process execution. A detailed description of these challenges associated with architectural dependencies is provided in Section \ref{subsubsec:dependencies}. 

\subsubsection{Strategies Across System Groups}
\label{subsubsec:grouping}
We now explore how different categories of systems balance the interplay between autonomy and alignment.
Based on our taxonomic classification and the resulting system profiles as illustrated in Fig. \ref{figure:radar-aspects}, we can categorize the selected 7 systems under analysis into three distinct system groups, which encompass general-purpose systems, central-controller systems, and role-agent systems. 
It's important to note that our categorization into these three groups, based on the systems chosen for this exploration, doesn't capture the entire spectrum of autonomous LLM-powered multi-agent systems. For a comprehensive overview of existing systems and system categories, we recommend referring to the recent surveys provided by \cite{wang2023survey, xi2023rise}. In the following, the key characteristics as observed from the corresponding system profiles are discussed. 

\begin{itemize}
    \item \textbf{General-Purpose Systems} - representing multi-agent systems designed for and adaptable to a broad spectrum of tasks and applications. Within the analyzed set of multi-agent systems, the following fall into this group: \textsc{Auto-GPT} \cite{autogpt2023}, \textsc{BabyAGI} \cite{babyagi2023}, \textsc{SuperAGI} \cite{superagi2023}, and \textsc{AgentGPT} \cite{agentgpt2023}. Goals are decomposed autonomously and represented as prioritized task lists (L2 \texttt{Decom}). They employ a multi-cycle process framework performed by dedicated task-management agents represented by certain generic agent types, including a single task-execution agent (see Section \ref{subsec:aa-ontology}). Relations and communications between these agents are strictly predefined, and agent conversations express as a monologue of the task-execution agent, resulting in low autonomy levels (L0) for communication protocol (\texttt{CommP}), and network management (\texttt{NetM}). The task-related actions are performed autonomously by the task-execution agent (mostly L2 autonomy \texttt{ActM}). while resource integration is based on provided mechanisms (\texttt{Integ}), the resources are selected and utilized by the LLM-powered in a self-organizing manner (L2 autonomy for \texttt{Util}), except for \textsc{CAMEL}; resulting in similar autonomy profiles for the aforementioned aspects.
    Besides from these commonalities, these systems distinguish in certain characteristics. Both \textsc{Auto-GPT} and \textsc{BabyAGI} employ generic task-execution agent, and provide no further alignment options at all. Moreover, these systems employ a generic task-execution agent with predefined agent roles and relations, reulting in L0 autonomy for \texttt{AGen} and \texttt{NetM}.
    In contrast, \textsc{SuperAGI} and \textsc{AgentGPT} employ execution agents with self-organizing agent roles (L2 autonomy for \texttt{RoleD}), an adaptable orchestration process (L1 for \texttt{Orch}), and some alignment options, especially for agent-specific aspects.
    Moreover, these systems employ execution agents, whose roles can be customized by the user (L1 alignment for \texttt{AGen}).

    \item \textbf{Central LLM Controller} - marks a third group specialized in leveraging and combining contextual resources for accomplishing the complex goals. \textsc{HuggingGPT} \cite{shen2023hugginggpt} serves as an archetype of such systems, utilizing resources especially in terms of existing ML models integrated via \textsc{Hugging Face}. 
    As already detailed in Section \ref{subsec:comp-classification}, \textsc{HuggingGPT} is characterized by a single central LLM-powered control agent with monologue-based reflection and planning. Language in terms of agent prompts as generic interface to manage the interplay between multiple specialized foundation models. In comparison to other systems or system groups, we see the highest levels of autonomy granted to this central agent (mostly L2); also see Fig.\ \ref{figure:radar-aspects} (d). 
    Furthermore, we see a finite and artefact-oriented process adaptable by the LLM-powered agent for orchestrating the different model-related tasks (L1 autonomy). As already stated above, beyond prompting the task, there are no further user-centric alignment options (L0 alignment).

    \item \textbf{Role-Agent Systems} - employ an interplay or simulation between multiple dedicated roles agents. This collaboration can serve different purposes, such as simulating a discussion or solving tasks that demand for a multi-perspective collaboration. With defined roles in a certain environment (such as in a software development project), their application is bound to this application domain or special purpose. Among the analyzed systems, \textsc{MetaGPT} \cite{hong2023metagpt} and \textsc{CAMEL} \cite{li2023camel} represent such systems. In contrast to the general-purpose systems, the execution agents play roles with dedicated responsibilities in a certain application domain. Furthermore, these role agents actually collaborate directly with each other. In case of the two exemplary systems, this collaboration is realized by communication protocols employing a dynamic exchange between agents with instructor and executor roles. 
    In particular, \textsc{CAMEL} employs two such role agents based on predefined agent types, but adjustable by the user. In ongoing strict dialogue cycles, the AI-user role agents instructs the AI-assistant role agent to execute the tasks (L0 autonomy for \texttt{CommP}). Similar to \textsc{SuperAGI}, \textsc{CAMEL} requires the user to specify the agents' roles (L1 alignment). \textsc{MetaGPT}, in contrast, internally assigns predefined roles with responsibilities alongside a waterfall development process (L0 alignment); thus, also expressing a finite and artefact-oriented process (L0 autonomy for \texttt{Orch}), terminating with the produced and tested software program. However, like in real-world software project, refinement iterations can follow, optional feedback cycles make it adaptable for the agents (L1 autonomy for \texttt{CommP}).
    
\end{itemize}
\textbf{Strategy Assessment.} Beyond differences in the applied communication protocols, it is the flexibility of agent roles (in relation to both autonomy and alignment) and further customization options for agent-specific aspects that distinguishes the systems' strategies (see above). 
However, when examining how the systems deal with autonomy and alignment across further aspects, most systems and system groups show similar strategies. 
The reasoning capabilities of LLM-powered agents are especially leveraged in areas demanding high autonomy, such as the goal decomposition, the actual execution of task-related actions, and the utilization of contextual resources. Interestingly, these high-autonomy aspects are mostly combined with low alignment levels, resulting in \textit{bounded autonomy} aspects (refer to Table \ref{table:alignment_autonomy}). 
A closer look at aspect interdependencies, as depicted in Fig.\ \ref{figure:feature-diagram}, reveals that these internally \textit{unbalanced} aspects are accompanied by other low-autonomy aspects equipped with limited flexibility, as follows:
\begin{compactitem}
    \item Autonomous decomposition directly depends on the user-prompted goal.
    \item Autonomous action management depends on strict or predefined communication protocol.
    \item Autonomous resource utilization depends on strict or predefined resource integration. 
\end{compactitem}

In these cases, the predefined and rule-based mechanisms serve as integrated alignment guiding and controlling the accurate operation of the dependent autonomous aspects. 

Based on the findings of the taxonomic classification, in the next Section \ref{sec:discussion}, we discuss challenges for current systems and reflect on the taxonomy's limitations and potentials.

\section{Discussion}
\label{sec:discussion}

This paper introduces and applies a novel comprehensive taxonomy, shedding light on the ways autonomous LLM-powered multi-agent systems manage the dynamic interplay between autonomy and alignment within their architectures.
When interpreted through the lens of our taxonomy, we encounter challenges and development potentials for current LLM-powered multi-agent systems, which are discussed in Section \ref{subsec:dis-sys}.
Moreover, in Section \ref{subsec:dis-tax}, we reflect on limitations and further potentials of the taxonomy itself.

\subsection{Challenges for Current Systems}
\label{subsec:dis-sys}

Our analysis of architectural dynamics inherent to current LLM-powered multi-agent systems, as detailed in Section \ref{sec:comparison}, reveals a number of challenges regarding the interplay between autonomy and alignment. In accordance with \cite{wang2023survey}, we recognize challenges related to the adaptability of agent collaboration. Moreover, our exploration indicates potentials for user-centric alignment options and controlling high-autonomy aspects.

\textbf{Agent Collaboration.} Among the systems analyzed, we especially observe limitations regarding collaboration modes and role-playing capabilities, as well as risks tied to prompt-driven collaboration techniques. 
\begin{itemize}
    \item \textbf{Adaptability of Communication Protocols:} 
    As discussed in Section \ref{subsec:comp-interpretation}, the collaboration between agents is mainly characterized by restricted communication protocols between predefined task-execution agents, such as instructor-and-executor relationships, or sequential or multi-cycle processes with predefined execution chains. Employing LLM-powered agents to manage and adapt the constellation of the agent network as well as their collaboration modes could pave the way for more creative problem-solving methods in task execution. 
    
\item \textbf{Dynamic Role-Playing:} In particular, we also see development potentials via the flexible collaboration between self-organizing role agents, such as for simulating the complex interplay within a certain application domain. As far as observable, the potential of engaging multiple perspectives through different roles and standpoints has not yet been fully sounded.

\item \textbf{Robustness of Prompt-driven Collaboration:} 
Collaboration between LLM-powered agents basically relies on prompt-driven message exchange, such as by delegating tasks, asking questions, or evaluating task results. This communication mechanism, founded on a sequence of prompts, heavily relies on the quality of LLM responses, which are susceptible to errors in terms of incorrect or hallucinated results \cite{maynez2020faithfulness,ji2023survey}.
However, without the integration of comprehensive and robust control mechanisms to check the quality of these responses, the system is vulnerable to inaccuracies, misunderstandings, and inefficiencies \cite{hong2023metagpt}.

\end{itemize}

\textbf{User-Centric Alignment.}
Within the scope of analyzed systems, user-centric alignment options are very rare. Alignment mechanisms are predominantly integrated into the system architecture (see Section \ref{subsec:comp-interpretation}). Drawing from this limitation, we see potentials in certain user-guided and real-time responsive alignment options. 

\begin{itemize}
    \item \textbf{User-Guided Alignment Options:} The options for users to access and influence the internal workings of the system are very limited. The internal composition and collaboration of the agents are mostly opaque to the user, which reduces transparency of the system operation. Exception to this represents the runtime documentation of agents' reflection and planning, provided by certain systems (see Section \ref{subsec:comp-classification}). 
    The customization of internal mechanisms is mostly not provided to users. Besides agent generation and role definition (offered by a few systems), there is potential for user modifications related to communication protocols, task orchestration, or result synthesis. 
    Corresponding to the aspect adaptability for LLM-powered agents (see above), modifying these internal mechanisms would enable the user in exploring alternative problem-solving ways. 

    \item \textbf{Real-Time Responsiveness:} 
    The obvious lack of real-time adjustment capabilities can be seen founded in the nature of autonomous agent systems, which is accomplishing the user-prompted goal without further human intervention. 
    However, as elaborated on in Section \ref{sec:taxonomy}, autonomy and alignment can be understood as complementary aspects. The absence of user interaction and control during runtime restricts the potential for dynamic alignment, thereby limiting the system's flexibility in response to changes in the operational context. 
    As detailed in Section \ref{subsubsec:alignment}, the interaction layer allows the integration of interceptor mechanisms. This not only allows real-time monitoring, addressing key concerns of explainable AI \cite{ribeiro2016should,xu2019explainable}, but also to implement effective feedback and intervention options \cite{laplante2004real,hellerstein2004feedback}.     
    Collaborative environments fostering \textit{hybrid teamwork}, comprising autonomous agents (or agents systems) and human co-workers are essentially built upon such real-time responsiveness, ensuring dynamic realignment while working towards shared goals \cite{khosla1997engineering,neef2006taxonomy,wei2023multi}.
  
\end{itemize}

\textbf{Controlling High-Autonomy Aspects.}
Besides prompting-related flaws such as inaccurate or hallucinated responses (see above), our engagement with the analyzed multi-agent systems has revealed additional operational issues.
Occasionally, we witness non-terminating activities, where the system falls into infinite loops. 
For instance, this can manifest via solutions continually fine-tuned under the premise of improvement, or the system operation is stuck in a never ending dialogue between two LLM-powered agents.
Conversely, system operations might terminate in a dead end when encountering a task that requires competencies or resources that are either unavailable or inaccessible.
Obviously, the corresponding control mechanisms (\textit{integrated alignment}) applied in such systems are ill-equipped to efficiently catch these kinds of exceptions. This insufficiency proves particularly concerning, as it undermines the reliability and effectiveness of these systems.
However, besides this symptomatic treatment, the reasons for these problems can be seen founded in architectural complexities, such as high-autonomy levels not adequately aligned or intertwined dependencies resulting from varying levels of autonomy (refer to Section \ref{subsubsec:dependencies}). 

\subsection{Limitations and Potentials of the Taxonomy}
\label{subsec:dis-tax}

For engineering the taxonomic system, we chose a pragmatic and technical perspective (see Section \ref{sec:taxonomy}) and explored its utility by the exemplary classification of seven selected LLM-powered multi-agent systems (see Section \ref{sec:comparison}). However, departing from this exploration, certain limitations and further potentials become evident. 

\textbf{Taxonomic System.}
Our taxonomy conceptualizes autonomy and alignment not as binary extremes in a one-dimensional continuum, but as interacting and synergistic aspects. This distinctions allows forming a two-dimensional matrix (see Section \ref{sec:taxonomy}) combining hierarchic levels of autonomy (from automated mechanisms to self-organizing agents) and alignment (from system-integrated to real-time responsive). This structure reflects the aforementioned triadic relationship between the key decision-making entities in the system (i.e., human users, rules and mechanisms, as well as LLM-powered agents) and their dynamic interplay (i.e., alignment, system operation, and collaboration), as illustrated in Fig.\ \ref{figure:triadic-relationship}. Augmenting this, we map this matrix onto different characteristic aspects derived from four applied architectural viewpoints (see Section \ref{subsec:tax-viewpoints}). 

\begin{itemize}
    \item \textbf{Autonomy Scope:}
Within this, we reference high autonomy to the agents' self-organization capabilities for decision-making and further operational impact (see Section \ref{subsubsec:autonomy}). However, it's essential to consider that autonomy can span beyond this definition, encompassing facets like an agent's ability for self-enhancement and proactive agency.
    \item \textbf{Alignment Scope:}
In turn, the alignment dimension employed by the taxonomy reflects two key aspects, i.e., the \textit{origin} of the alignment, and the \textit{moment} of its communication to the system (see Section \ref{subsubsec:alignment}). In combination with the architectural dimension, we also reflect the \textit{architectural or functional scope} of the alignment technique in terms of the viewpoint-specific aspects.  
However, one must note that this dimension does not reflect the quality, efficacy, or depth of the applied techniques.
    \item \textbf{Scope of Architectural Aspects:}
As detailed in Section \ref{subsec:tax-matrices}, the taxonomy adopts 12 architectural aspects inherent to the four architectural viewpoints characteristic for LLM-powered multi-agent systems. The viewpoints are oriented to Kruchten's viewpoint model for software architecture \cite{kruchten19954+}, a recognized standard in this field. However, as there exist more viewpoint models reflecting further concerns and perspectives on software systems, there might also be further architectural aspects possibly relevant to autonomous LLM-powered multi-agent systems. Considering the ongoing evolution in the field, these adaptions become crucial.

\end{itemize}

\textbf{Expressiveness of Taxonomic Classification.}
The scope of the taxonomic structure forms the foundation for the taxonomy's analytical power enabling conclusions about the classified systems under analysis.

\begin{itemize}
    \item \textbf{Levels as Strengths and Weaknesses:}
    It is important to understand that higher levels in autonomy and alignment, termed as \textit{user-responsive autonomy} (see Table \ref{table:alignment_autonomy}), might not always be the optimal system configuration for every scenario. Indeed, high autonomy can deviate from the intended goal and therefore needs to be aligned accordingly. In certain situations, a system with modest autonomy could be considered the best choice. Given the intention to automate a repetitive set of routine tasks with predictable variables and contextual requirements, a static autonomy with predefined rules and mechanisms would not be just sufficient, but also provide a higher reliability. If there is no need to include user-specific information, a combination with an integrated alignment can be seen as best choice (\textit{rule-driven automation}). 

    \item \textbf{System Efficiency and Accuracy:}
    As previously elaborated, our taxonomy focuses on the architectural complexities driven by the dynamics between autonomy and alignment, rather than evaluating functional performance metrics like operational efficiency or accuracy. 
    Neither recent surveys in the field \cite{wang2023survey,ji2023survey} do measure the systems' performance, such as in terms of efficiency, accuracy, or scalability.
    However, while engaging with the analyzed systems, we observed substantial differences among them, reflecting the exploratory state and the ongoing rapid evolution of the domain. 
    For measuring their functional performance, benchmarks and methods could be adopted similar to those presented in \cite{bubeck2023sparks}. 

    \item \textbf{Balancing Techniques:}
    As reported in Section \ref{subsec:comp-interpretation}, we have identified different balancing strategies across the system architectures. In this context, it is important to notice that aspects marked as \textit{unbalanced} (for example, combining high-autonomy and low-alignment levels) might be actually controlled or balanced via automated mechanisms applied by another aspect (static autonomy and integrated alignment). Within the analyzed systems, user-centric alignment options are barely applied to curb the wildness of high-autonomy aspects. It would be interesting, to investigate and compare in detail, how integrated alignment techniques are employed to deal with the challenges and complexities of agent-driven autonomy. 
    
\end{itemize}

\textbf{Practical Implications.}
Drawing from the information value provided by the classification results, we can distinguish considerations regarding the practical utility and relevance of the taxonomy.

\begin{itemize}
    
    \item \textbf{Analysis Purposes:}
    The analysis and understanding of these dynamic architectural complexities can serve different purposes, such as: 
\begin{compactitem}
    \item Comparing, selecting, and applying available multi-agent systems in the context of given scenarios with certain requirements for autonomy and alignment.
    \item Reasoning about architectural design options for the development of novel multi-agent systems.
    \item Scrutinizing and rethinking strategies for balancing levels of autonomy and alignment.
    \item Building a foundational framework for additional analysis techniques or complementing them, such as measuring the functional system capabilities (see above).
\end{compactitem}

    \item \textbf{Ongoing Evolution:}
    As underscored by recent surveys \cite{wang2023survey,xi2023rise}, the field of autonomous LLM-powered multi-agent systems is characterized by an ongoing rapid evolution showcasing a dynamically growing number of approaches featuring diverse architectures and a wide spectrum of system-maturity levels. 
    While designed to abstract from concrete system specifics, the taxonomic system might need periodic updates to accommodate this dynamically evolving landscape.

    \item \textbf{Broader Applicability.}
    Tailored to address the characteristics of autonomous LLM-powered multi-agent architectures (refer to Section \ref{sec:architecture}), the foundational principles of our taxonomy, however, seem to be transferable to other AI systems. Certain segments of the taxonomic structure can be seen as universally applicable across AI architectures. Conversely, facets specifically tailored to multi-agent systems, such as the aspects inherent to the agent composition and multi-agent collaboration viewpoints, would require corresponding adjustments.
    
\end{itemize}

\section{Conclusion}
\label{sec:conclusion}

In this paper, we have introduced a comprehensive multi-dimensional taxonomy engineered to analyze how autonomous LLM-powered multi-agent systems balance the dynamic interplay between autonomy and alignment across their system architectures.
For this purpose, the taxonomy employs a matrix that combines hierarchical levels of autonomy and alignment. This matrix is then mapped onto various architectural aspects organized by four architectural viewpoints reflecting different complementary concerns and perspectives.
The resulting taxonomic system enables the assessment of interdependent aspect configurations in a wide spectrum, ranging from simple configurations, such as predefined mechanisms combined with system-integrated alignment techniques (\textit{rule-driven automation}), to sophisticated configurations, such as self-organizing agency responsive to user feedback and evolving conditions (\textit{user-responsive autonomy}). 
Applied to 12 distinct architectural aspects inherent to viewpoints, such as goal-driven task management, multi-agent collaboration, agent composition, and context interaction, this taxonomy allows for a nuanced analysis and understanding of architectural complexities within autonomous LLM-powered multi-agent systems.

Through our taxonomy's application to seven selected LLM-powered multi-agent systems, its practical relevance and utility has been illustrated.
In particular, it has been shown that a combined assessment of autonomy and alignment levels across the architectural aspects of each multi-agent system allows for identifying system profiles that can indicate certain strategies for balancing the dynamic interplay between autonomy and alignment.
This exploration of exemplary current systems also revealed several challenges. 
Most prominently, we observed a lack of user-centric alignment options across all systems, with little user-guided alignment, but no real-time responsive alignment at all. 
Moreover, the systems exhibit high autonomy levels mostly for certain aspects, such as the goal decomposition, the action management, or the utilization of contextual resources. 
In contrast, other key aspects of the system operation show limited autonomy; aspects such as managing the communication protocol, memory usage, or agent network are largely static, leaning heavily on predefined mechanisms.

Based on these and further findings, we especially see two promising avenues for the evolution of autonomous LLM-powered multi-agent systems.
Firstly, by employing adaptable and self-organizing communication protocols and agent networks, the systems' role-playing capabilities could be enhances, which enables them to better simulate complex multi-perspective environments. By reflecting and weighing up diverse standpoints and strategies, this could also pave the way for more in-depth inter-agent discussions and creativity in problem solving. 
Secondly, the exploration of real-time responsive systems, which can adapt to evolving conditions as well as to user feedback during runtime, would foster dynamic collaboration and hybrid teamwork between LLM-powered agents and human users.

Departing from an exploratory stage, the field of autonomous LLM-powered multi-agent systems is rapidly evolving, resulting in a growing number of promising approaches and innovative architectures. 
With their current capabilities and inherent potentials, such as multi-perspective domain simulations or collaborative environments of autonomous agents and human coworkers, these systems could significantly contribute to the progression towards advanced stages of artificial intelligence, such as AGI or ASI. 
From a pragmatic perspective, there are numerous opportunities for combining LLMs as general purpose technology with the specifics of various application domains. LLM-based multi-agent systems can serve as foundation for developing corresponding domain-specific application layers.
The architectural complexities resulting from the dynamic interplay between autonomy and alignment can be seen as one of the key challenges in such systems. 
By providing a systematic framework for analyzing these complexities, our taxonomy aims to contribute to these ongoing efforts.

For our subsequent endeavors, we aim at developing a comprehensive overview and comparison of existing autonomous LLM-powered multi-agent systems, complementing existing literature reviews in the field \cite{wang2023survey,xi2023rise}. 
To this end, we intend to analyze and classify available systems using our taxonomy. 
The identified system profiles and balancing strategies resulting from this analysis will then be combined with further investigations of functional system capabilities.
In addition, driven by the potentials identified during the taxonomic classification of selected systems, we currently explore the development of an LLM-powered multi-agent system that aims at combining high levels of agency with real-time user-centric control mechanisms.

Building on the foundation of our taxonomy, future initiatives could venture into the following areas:
A dedicated exploration, assessment, and systematization of alignment techniques, particularly tailored for LLM-based interaction and application layers, could serve as reference for future systems.
Moreover, the conception of a methodological framework with instruments and benchmarks for measuring the functional capabilities of LLM-powered multi-agent systems could provide a structured template to evaluate key metrics like efficiency, accuracy, and scalability of these systems.

\vfill

\section*{Acknowledgements}
The author gratefully acknowledges the support from the "Gesellschaft für Forschungsförderung (GFF)" of Lower Austria, as this research was conducted at Ferdinand Porsche Mobile University of Applied Sciences (FERNFH) as part of the "Digital Transformation Hub" project funded by the GFF.

\renewcommand{\refname}{References}
\bibliographystyle{abbrv}

\bibliography{references}

\begin{thebibliography}{100}

\bibitem{amodei2016concrete}
D.~Amodei, C.~Olah, J.~Steinhardt, P.~Christiano, J.~Schulman, and D.~Man{\'e}.
\newblock Concrete problems in {AI} safety.
\newblock {\em arXiv preprint arXiv:1606.06565}, 2016.

\bibitem{arrieta2020explainable}
A.~B. Arrieta, N.~D{\'\i}az-Rodr{\'\i}guez, J.~Del~Ser, A.~Bennetot, S.~Tabik,
  A.~Barbado, S.~Garc{\'\i}a, S.~Gil-L{\'o}pez, D.~Molina, R.~Benjamins, et~al.
\newblock Explainable artificial intelligence ({XAI}): Concepts, taxonomies,
  opportunities and challenges toward responsible {AI}.
\newblock {\em Information fusion}, 58:82--115, 2020.

\bibitem{askell2021general}
A.~Askell, Y.~Bai, A.~Chen, D.~Drain, D.~Ganguli, T.~Henighan, A.~Jones,
  N.~Joseph, B.~Mann, N.~DasSarma, et~al.
\newblock A general language assistant as a laboratory for alignment.
\newblock {\em arXiv preprint arXiv:2112.00861}, 2021.

\bibitem{bass2003software}
L.~Bass, P.~Clements, and R.~Kazman.
\newblock {\em Software architecture in practice}.
\newblock Addison-Wesley Professional, 2003.

\bibitem{batory2005feature}
D.~Batory.
\newblock Feature models, grammars, and propositional formulas.
\newblock In {\em 9th International Software Product Line Conference}, pages
  7--20, 2005.

\bibitem{beer2014toward}
J.~M. Beer, A.~D. Fisk, and W.~A. Rogers.
\newblock Toward a framework for levels of robot autonomy in human-robot
  interaction.
\newblock {\em Journal of human-robot interaction}, 3(2):74, 2014.

\bibitem{bird1993toward}
S.~D. Bird.
\newblock Toward a taxonomy of multi-agent systems.
\newblock {\em International Journal of Man-Machine Studies}, 39(4):689--704,
  1993.

\bibitem{boehm1988understanding}
B.~W. Boehm and P.~N. Papaccio.
\newblock Understanding and controlling software costs.
\newblock {\em IEEE transactions on software engineering}, 14(10):1462--1477,
  1988.

\bibitem{bommasani2021opportunities}
R.~Bommasani, D.~A. Hudson, E.~Adeli, R.~Altman, S.~Arora, S.~von Arx, M.~S.
  Bernstein, J.~Bohg, A.~Bosselut, E.~Brunskill, et~al.
\newblock On the opportunities and risks of foundation models.
\newblock {\em arXiv preprint arXiv:2108.07258}, 2021.

\bibitem{bostrom2017superintelligence}
N.~Bostrom.
\newblock {\em Superintelligence}.
\newblock Dunod, 2017.

\bibitem{brown2020language}
T.~Brown, B.~Mann, N.~Ryder, M.~Subbiah, J.~D. Kaplan, P.~Dhariwal,
  A.~Neelakantan, P.~Shyam, G.~Sastry, A.~Askell, et~al.
\newblock Language models are few-shot learners.
\newblock {\em Advances in neural information processing systems},
  33:1877--1901, 2020.

\bibitem{brustoloni1991autonomous}
J.~C. Brustoloni.
\newblock {\em Autonomous agents: Characterization and requirements}.
\newblock Carnegie Mellon University, 1991.

\bibitem{bubeck2023sparks}
S.~Bubeck, V.~Chandrasekaran, R.~Eldan, J.~Gehrke, E.~Horvitz, E.~Kamar,
  P.~Lee, Y.~T. Lee, Y.~Li, S.~Lundberg, et~al.
\newblock Sparks of artificial general intelligence: Early experiments with
  {GPT-4}.
\newblock {\em arXiv preprint arXiv:2303.12712}, 2023.

\bibitem{langchain2022}
H.~Chase.
\newblock {LangChain}.
\newblock \url{https://github.com/langchain-ai/langchain}, 2022.

\bibitem{chowdhery2022palm}
A.~Chowdhery, S.~Narang, J.~Devlin, M.~Bosma, G.~Mishra, A.~Roberts, P.~Barham,
  H.~W. Chung, C.~Sutton, S.~Gehrmann, et~al.
\newblock Palm: Scaling language modeling with pathways.
\newblock {\em arXiv preprint arXiv:2204.02311}, 2022.

\bibitem{clements2003documenting}
P.~Clements, D.~Garlan, R.~Little, R.~Nord, and J.~Stafford.
\newblock Documenting software architectures: views and beyond.
\newblock In {\em 25th International Conference on Software Engineering, 2003.
  Proceedings.}, pages 740--741. IEEE, 2003.

\bibitem{csepregi2023effect}
L.~M. Csepregi.
\newblock The effect of context-aware {LLM}-based {NPC} conversations on player
  engagement in role-playing video games.
\newblock 2023.

\bibitem{du2023improving}
Y.~Du, S.~Li, A.~Torralba, J.~B. Tenenbaum, and I.~Mordatch.
\newblock Improving factuality and reasoning in language models through
  multiagent debate.
\newblock {\em arXiv preprint arXiv:2305.14325}, 2023.

\bibitem{dudek1996taxonomy}
G.~Dudek, M.~R. Jenkin, E.~Milios, and D.~Wilkes.
\newblock A taxonomy for multi-agent robotics.
\newblock {\em Autonomous Robots}, 3:375--397, 1996.

\bibitem{fabiano2023fast}
F.~Fabiano, V.~Pallagani, M.~B. Ganapini, L.~Horesh, A.~Loreggia, K.~Murugesan,
  F.~Rossi, and B.~Srivastava.
\newblock Fast and slow planning.
\newblock {\em arXiv preprint arXiv:2303.04283}, 2023.

\bibitem{franklin1996agent}
S.~Franklin and A.~Graesser.
\newblock Is it an agent, or just a program?: A taxonomy for autonomous agents.
\newblock In {\em International workshop on agent theories, architectures, and
  languages}, pages 21--35. Springer, 1996.

\bibitem{gao2023s}
C.~Gao, X.~Lan, Z.~Lu, J.~Mao, J.~Piao, H.~Wang, D.~Jin, and Y.~Li.
\newblock S\textsuperscript{3}: Social-network simulation system with large
  language model-empowered agents.
\newblock {\em arXiv preprint arXiv:2307.14984}, 2023.

\bibitem{gruber1995toward}
T.~R. Gruber.
\newblock Toward principles for the design of ontologies used for knowledge
  sharing?
\newblock {\em International journal of human-computer studies},
  43(5-6):907--928, 1995.

\bibitem{guizzardi2002general}
G.~Guizzardi, H.~Herre, and G.~Wagner.
\newblock On the general ontological foundations of conceptual modeling.
\newblock In {\em Conceptual Modeling—ER 2002: 21st International Conference
  on Conceptual Modeling Tampere, Finland, October 7--11, 2002 Proceedings 21},
  pages 65--78. Springer, 2002.

\bibitem{haendler2020ontology}
T.~Haendler and G.~Neumann.
\newblock Ontology-based analysis and design of educational games for software
  refactoring.
\newblock In {\em Computer Supported Education: 11th International Conference,
  CSEDU 2019, Heraklion, Crete, Greece, May 2-4, 2019, Revised Selected
  Papers}, pages 602--628. Springer, 2020.

\bibitem{hao2023chatllm}
R.~Hao, L.~Hu, W.~Qi, Q.~Wu, Y.~Zhang, and L.~Nie.
\newblock {ChatLLM} network: More brains, more intelligence.
\newblock {\em arXiv preprint arXiv:2304.12998}, 2023.

\bibitem{hellerstein2004feedback}
J.~L. Hellerstein, Y.~Diao, S.~Parekh, and D.~M. Tilbury.
\newblock {\em Feedback control of computing systems}.
\newblock John Wiley \& Sons, 2004.

\bibitem{hong2023metagpt}
S.~Hong, X.~Zheng, J.~Chen, Y.~Cheng, C.~Zhang, Z.~Wang, S.~K.~S. Yau, Z.~Lin,
  L.~Zhou, C.~Ran, et~al.
\newblock {MetaGPT}: Meta programming for multi-agent collaborative framework.
\newblock {\em arXiv preprint arXiv:2308.00352}, 2023.

\bibitem{sae2016taxonomy}
S.~International.
\newblock Taxonomy and definitions for terms related to driving automation
  systems for on-road motor vehicles, 2016.

\bibitem{ji2023survey}
Z.~Ji, N.~Lee, R.~Frieske, T.~Yu, D.~Su, Y.~Xu, E.~Ishii, Y.~J. Bang,
  A.~Madotto, and P.~Fung.
\newblock Survey of hallucination in natural language generation.
\newblock {\em ACM Computing Surveys}, 55(12):1--38, 2023.

\bibitem{johnson2019billion}
J.~Johnson, M.~Douze, and H.~J{\'e}gou.
\newblock Billion-scale similarity search with {GPUs}.
\newblock {\em IEEE Transactions on Big Data}, 7(3):535--547, 2019.

\bibitem{kaddour2023challenges}
J.~Kaddour, J.~Harris, M.~Mozes, H.~Bradley, R.~Raileanu, and R.~McHardy.
\newblock Challenges and applications of large language models.
\newblock {\em arXiv preprint arXiv:2307.10169}, 2023.

\bibitem{kahneman2011thinking}
D.~Kahneman.
\newblock {\em Thinking, fast and slow}.
\newblock Macmillan, 2011.

\bibitem{khosla1997engineering}
R.~Khosla.
\newblock {\em Engineering intelligent hybrid multi-agent systems}.
\newblock Springer Science \& Business Media, 1997.

\bibitem{kiczales1997aspect}
G.~Kiczales, J.~Lamping, A.~Mendhekar, C.~Maeda, C.~Lopes, J.-M. Loingtier, and
  J.~Irwin.
\newblock Aspect-oriented programming.
\newblock In {\em ECOOP'97—Object-Oriented Programming: 11th European
  Conference Jyv{\"a}skyl{\"a}, Finland, June 9--13, 1997 Proceedings 11},
  pages 220--242. Springer, 1997.

\bibitem{kitchenham1999towards}
B.~A. Kitchenham, G.~H. Travassos, A.~Von~Mayrhauser, F.~Niessink, N.~F.
  Schneidewind, J.~Singer, S.~Takada, R.~Vehvilainen, and H.~Yang.
\newblock Towards an ontology of software maintenance.
\newblock {\em Journal of Software Maintenance: Research and Practice},
  11(6):365--389, 1999.

\bibitem{kojima2022large}
T.~Kojima, S.~S. Gu, M.~Reid, Y.~Matsuo, and Y.~Iwasawa.
\newblock Large language models are zero-shot reasoners.
\newblock {\em Advances in neural information processing systems},
  35:22199--22213, 2022.

\bibitem{kruchten19954+}
P.~B. Kruchten.
\newblock Architectural blueprints — the “4+1” view model of software
  architecture.
\newblock {\em IEEE software}, 12(6):42--50, 1995.

\bibitem{labrou1998semantics}
Y.~Labrou and T.~Finin.
\newblock Semantics and conversations for an agent communication language.
\newblock {\em arXiv preprint cs/9809034}, 1998.

\bibitem{laplante2004real}
P.~A. Laplante et~al.
\newblock {\em Real-time systems design and analysis}.
\newblock Wiley New York, 2004.

\bibitem{li2023camel}
G.~Li, H.~A. A.~K. Hammoud, H.~Itani, D.~Khizbullin, and B.~Ghanem.
\newblock {CAMEL}: Communicative agents for "mind" exploration of large scale
  language model society.
\newblock {\em arXiv preprint arXiv:2303.17760}, 2023.

\bibitem{liang2023encouraging}
T.~Liang, Z.~He, W.~Jiao, X.~Wang, Y.~Wang, R.~Wang, Y.~Yang, Z.~Tu, and
  S.~Shi.
\newblock Encouraging divergent thinking in large language models through
  multi-agent debate.
\newblock {\em arXiv preprint arXiv:2305.19118}, 2023.

\bibitem{lin2023swiftsage}
B.~Y. Lin, Y.~Fu, K.~Yang, P.~Ammanabrolu, F.~Brahman, S.~Huang,
  C.~Bhagavatula, Y.~Choi, and X.~Ren.
\newblock {SwiftSage}: A generative agent with fast and slow thinking for
  complex interactive tasks.
\newblock {\em arXiv preprint arXiv:2305.17390}, 2023.

\bibitem{maes1995artificial}
P.~Maes.
\newblock Artificial life meets entertainment: lifelike autonomous agents.
\newblock {\em Communications of the ACM}, 38(11):108--114, 1995.

\bibitem{maynez2020faithfulness}
J.~Maynez, S.~Narayan, B.~Bohnet, and R.~McDonald.
\newblock On faithfulness and factuality in abstractive summarization.
\newblock {\em arXiv preprint arXiv:2005.00661}, 2020.

\bibitem{meyer1992applying}
B.~Meyer.
\newblock Applying'design by contract'.
\newblock {\em Computer}, 25(10):40--51, 1992.

\bibitem{mikolov2013efficient}
T.~Mikolov, K.~Chen, G.~Corrado, and J.~Dean.
\newblock Efficient estimation of word representations in vector space.
\newblock {\em arXiv preprint arXiv:1301.3781}, 2013.

\bibitem{minsky1988society}
M.~Minsky.
\newblock {\em The Society of mind}.
\newblock Simon and Schuster, 1988.

\bibitem{mintzberg1989structuring}
H.~Mintzberg.
\newblock {\em The structuring of organizations}.
\newblock Springer, 1989.

\bibitem{moya2007towards}
L.~J. Moya and A.~Tolk.
\newblock Towards a taxonomy of agents and multi-agent systems.
\newblock In {\em SpringSim (2)}, pages 11--18, 2007.

\bibitem{babyagi2023}
Y.~Nakajima.
\newblock {BabyAGI}.
\newblock \url{https://github.com/yoheinakajima/babyagi}, 2023.

\bibitem{narendra2012stable}
K.~S. Narendra and A.~M. Annaswamy.
\newblock {\em Stable adaptive systems}.
\newblock Courier Corporation, 2012.

\bibitem{naveed2023comprehensive}
H.~Naveed, A.~U. Khan, S.~Qiu, M.~Saqib, S.~Anwar, M.~Usman, N.~Barnes, and
  A.~Mian.
\newblock A comprehensive overview of large language models.
\newblock {\em arXiv preprint arXiv:2307.06435}, 2023.

\bibitem{neef2006taxonomy}
M.~Neef.
\newblock A taxonomy of human-agent team collaborations.
\newblock In {\em Proceedings of the 18th BeNeLux Conference on Artificial
  Intelligence (BNAIC 2006)}, pages 245--250, 2006.

\bibitem{UML}
{Object Management Group}.
\newblock {Unified Modeling Language} -- version 2.5.1.
\newblock \url{https://www.omg.org/spec/UML/2.5.1}, Dec. 2017.

\bibitem{ouyang2022training}
L.~Ouyang, J.~Wu, X.~Jiang, D.~Almeida, C.~Wainwright, P.~Mishkin, C.~Zhang,
  S.~Agarwal, K.~Slama, A.~Ray, et~al.
\newblock Training language models to follow instructions with human feedback.
\newblock {\em Advances in Neural Information Processing Systems},
  35:27730--27744, 2022.

\bibitem{o2008ambidexterity}
C.~A. O’reilly~Iii and M.~L. Tushman.
\newblock Ambidexterity as a dynamic capability: Resolving the innovator's
  dilemma.
\newblock {\em Research in organizational behavior}, 28:185--206, 2008.

\bibitem{parasuraman2000model}
R.~Parasuraman, T.~B. Sheridan, and C.~D. Wickens.
\newblock A model for types and levels of human interaction with automation.
\newblock {\em IEEE Transactions on systems, man, and cybernetics-Part A:
  Systems and Humans}, 30(3):286--297, 2000.

\bibitem{park2023generative}
J.~S. Park, J.~C. O'Brien, C.~J. Cai, M.~R. Morris, P.~Liang, and M.~S.
  Bernstein.
\newblock Generative agents: Interactive simulacra of human behavior.
\newblock {\em arXiv preprint arXiv:2304.03442}, 2023.

\bibitem{patil2023gorilla}
S.~G. Patil, T.~Zhang, X.~Wang, and J.~E. Gonzalez.
\newblock Gorilla: Large language model connected with massive {APIs}.
\newblock {\em arXiv preprint arXiv:2305.15334}, 2023.

\bibitem{qian2023communicative}
C.~Qian, X.~Cong, C.~Yang, W.~Chen, Y.~Su, J.~Xu, Z.~Liu, and M.~Sun.
\newblock Communicative agents for software development.
\newblock {\em arXiv preprint arXiv:2307.07924}, 2023.

\bibitem{rahmati2017ifttt}
A.~Rahmati, E.~Fernandes, J.~Jung, and A.~Prakash.
\newblock {IFTTT vs. Zapier}: A comparative study of trigger-action programming
  frameworks.
\newblock {\em arXiv preprint arXiv:1709.02788}, 2017.

\bibitem{ribeiro2016should}
M.~T. Ribeiro, S.~Singh, and C.~Guestrin.
\newblock "why should i trust you?" explaining the predictions of any
  classifier.
\newblock In {\em Proceedings of the 22nd ACM SIGKDD international conference
  on knowledge discovery and data mining}, pages 1135--1144, 2016.

\bibitem{rozanski2012software}
N.~Rozanski and E.~Woods.
\newblock {\em Software systems architecture: working with stakeholders using
  viewpoints and perspectives}.
\newblock Addison-Wesley, 2012.

\bibitem{russell2019human}
S.~Russell.
\newblock {\em Human compatible: Artificial intelligence and the problem of
  control}.
\newblock Penguin, 2019.

\bibitem{russell2022artificial}
S.~Russell.
\newblock Artificial intelligence and the problem of control.
\newblock {\em Perspectives on Digital Humanism}, page~19, 2022.

\bibitem{russell2015research}
S.~Russell, D.~Dewey, and M.~Tegmark.
\newblock Research priorities for robust and beneficial artificial
  intelligence.
\newblock {\em AI magazine}, 36(4):105--114, 2015.

\bibitem{santu2023teler}
S.~K.~K. Santu and D.~Feng.
\newblock {TELeR}: A general taxonomy of {LLM} prompts for benchmarking complex
  tasks.
\newblock {\em arXiv preprint arXiv:2305.11430}, 2023.

\bibitem{schobbens2007generic}
P.-Y. Schobbens, P.~Heymans, J.-C. Trigaux, and Y.~Bontemps.
\newblock Generic semantics of feature diagrams.
\newblock {\em Computer networks}, 51(2):456--479, 2007.

\bibitem{shen2023hugginggpt}
Y.~Shen, K.~Song, X.~Tan, D.~Li, W.~Lu, and Y.~Zhuang.
\newblock {HuggingGPT}: Solving {AI} tasks with {ChatGPT} and its friends in
  {Hugging Face}.
\newblock {\em arXiv preprint arXiv:2303.17580}, 2023.

\bibitem{agentgpt2023}
A.~Shrestha, S.~Subedi, and A.~Watkins.
\newblock {AgentGPT}.
\newblock \url{https://github.com/reworkd/AgentGPT}, 2023.

\bibitem{shum2023automatic}
K.~Shum, S.~Diao, and T.~Zhang.
\newblock Automatic prompt augmentation and selection with chain-of-thought
  from labeled data.
\newblock {\em arXiv preprint arXiv:2302.12822}, 2023.

\bibitem{singh1998agent}
M.~P. Singh.
\newblock Agent communication languages: Rethinking the principles.
\newblock {\em Computer}, 31(12):40--47, 1998.

\bibitem{sloman1996empirical}
S.~A. Sloman.
\newblock The empirical case for two systems of reasoning.
\newblock {\em Psychological bulletin}, 119(1):3, 1996.

\bibitem{sowa1995top}
J.~F. Sowa.
\newblock Top-level ontological categories.
\newblock {\em International journal of human-computer studies},
  43(5-6):669--685, 1995.

\bibitem{thoppilan2022lamda}
R.~Thoppilan, D.~De~Freitas, J.~Hall, N.~Shazeer, A.~Kulshreshtha, H.-T. Cheng,
  A.~Jin, T.~Bos, L.~Baker, Y.~Du, et~al.
\newblock Lamda: Language models for dialog applications.
\newblock {\em arXiv preprint arXiv:2201.08239}, 2022.

\bibitem{autogpt2023}
{Torantulino et al.}
\newblock {Auto-GPT}.
\newblock \url{https://github.com/Significant-Gravitas/Auto-GPT}, 2023.

\bibitem{tosic2004towards}
P.~T. Tosic and G.~A. Agha.
\newblock Towards a hierarchical taxonomy of autonomous agents.
\newblock In {\em 2004 IEEE International Conference on Systems, Man and
  Cybernetics (IEEE Cat. No. 04CH37583)}, volume~4, pages 3421--3426. IEEE,
  2004.

\bibitem{superagi2023}
{TransformerOptimus et al.}
\newblock {SuperAGI}.
\newblock \url{https://github.com/TransformerOptimus/SuperAGI}, 2023.

\bibitem{tufte2001visual}
E.~R. Tufte.
\newblock {\em The visual display of quantitative information}, volume~2.
\newblock Graphics press Cheshire, CT, 2001.

\bibitem{usman2017taxonomies}
M.~Usman, R.~Britto, J.~B{\"o}rstler, and E.~Mendes.
\newblock Taxonomies in software engineering: A systematic mapping study and a
  revised taxonomy development method.
\newblock {\em Information and Software Technology}, 85:43--59, 2017.

\bibitem{van2004design}
H.~Van Dyke~Parunak, S.~Brueckner, M.~Fleischer, and J.~Odell.
\newblock A design taxonomy of multi-agent interactions.
\newblock In {\em Agent-Oriented Software Engineering IV: 4th
  InternationalWorkshop, AOSE 2003, Melbourne, Australia, July 15, 2003.
  Revised Papers 4}, pages 123--137. Springer, 2004.

\bibitem{wang2023voyager}
G.~Wang, Y.~Xie, Y.~Jiang, A.~Mandlekar, C.~Xiao, Y.~Zhu, L.~Fan, and
  A.~Anandkumar.
\newblock Voyager: An open-ended embodied agent with large language models.
\newblock {\em arXiv preprint arXiv:2305.16291}, 2023.

\bibitem{wang2023survey}
L.~Wang, C.~Ma, X.~Feng, Z.~Zhang, H.~Yang, J.~Zhang, Z.~Chen, J.~Tang,
  X.~Chen, Y.~Lin, et~al.
\newblock A survey on large language model based autonomous agents.
\newblock {\em arXiv preprint arXiv:2308.11432}, 2023.

\bibitem{wang2023plan}
L.~Wang, W.~Xu, Y.~Lan, Z.~Hu, Y.~Lan, R.~K.-W. Lee, and E.-P. Lim.
\newblock Plan-and-solve prompting: Improving zero-shot chain-of-thought
  reasoning by large language models.
\newblock {\em arXiv preprint arXiv:2305.04091}, 2023.

\bibitem{wang2023recursively}
Q.~Wang, L.~Ding, Y.~Cao, Z.~Tian, S.~Wang, D.~Tao, and L.~Guo.
\newblock Recursively summarizing enables long-term dialogue memory in large
  language models.
\newblock {\em arXiv preprint arXiv:2308.15022}, 2023.

\bibitem{wang2023augmenting}
W.~Wang, L.~Dong, H.~Cheng, X.~Liu, X.~Yan, J.~Gao, and F.~Wei.
\newblock Augmenting language models with long-term memory.
\newblock {\em arXiv preprint arXiv:2306.07174}, 2023.

\bibitem{wang2023unleashing}
Z.~Wang, S.~Mao, W.~Wu, T.~Ge, F.~Wei, and H.~Ji.
\newblock Unleashing cognitive synergy in large language models: A task-solving
  agent through multi-persona self-collaboration.
\newblock {\em arXiv preprint arXiv:2307.05300}, 2023.

\bibitem{wei2023multi}
J.~Wei, K.~Shuster, A.~Szlam, J.~Weston, J.~Urbanek, and M.~Komeili.
\newblock Multi-party chat: Conversational agents in group settings with humans
  and models.
\newblock {\em arXiv preprint arXiv:2304.13835}, 2023.

\bibitem{wei2022chain}
J.~Wei, X.~Wang, D.~Schuurmans, M.~Bosma, F.~Xia, E.~Chi, Q.~V. Le, D.~Zhou,
  et~al.
\newblock Chain-of-thought prompting elicits reasoning in large language
  models.
\newblock {\em Advances in Neural Information Processing Systems},
  35:24824--24837, 2022.

\bibitem{white2023prompt}
J.~White, Q.~Fu, S.~Hays, M.~Sandborn, C.~Olea, H.~Gilbert, A.~Elnashar,
  J.~Spencer-Smith, and D.~C. Schmidt.
\newblock A prompt pattern catalog to enhance prompt engineering with
  {ChatGPT}.
\newblock {\em arXiv preprint arXiv:2302.11382}, 2023.

\bibitem{wolf2023fundamental}
Y.~Wolf, N.~Wies, Y.~Levine, and A.~Shashua.
\newblock Fundamental limitations of alignment in large language models.
\newblock {\em arXiv preprint arXiv:2304.11082}, 2023.

\bibitem{wooldridge2009introduction}
M.~Wooldridge.
\newblock {\em An introduction to multiagent systems}.
\newblock John wiley \& sons, 2009.

\bibitem{wooldridge1995intelligent}
M.~Wooldridge and N.~R. Jennings.
\newblock Intelligent agents: Theory and practice.
\newblock {\em The knowledge engineering review}, 10(2):115--152, 1995.

\bibitem{xi2023rise}
Z.~Xi, W.~Chen, X.~Guo, W.~He, Y.~Ding, B.~Hong, M.~Zhang, J.~Wang, S.~Jin,
  E.~Zhou, et~al.
\newblock The rise and potential of large language model based agents: A
  survey.
\newblock {\em arXiv preprint arXiv:2309.07864}, 2023.

\bibitem{xu2019explainable}
F.~Xu, H.~Uszkoreit, Y.~Du, W.~Fan, D.~Zhao, and J.~Zhu.
\newblock Explainable {AI}: A brief survey on history, research areas,
  approaches and challenges.
\newblock In {\em Natural Language Processing and Chinese Computing: 8th CCF
  International Conference, NLPCC 2019, Dunhuang, China, October 9--14, 2019,
  Proceedings, Part II 8}, pages 563--574. Springer, 2019.

\bibitem{yudkowsky2016ai}
E.~Yudkowsky.
\newblock The {AI} alignment problem: why it is hard, and where to start.
\newblock {\em Symbolic Systems Distinguished Speaker}, 4, 2016.

\bibitem{zhang2022opt}
S.~Zhang, S.~Roller, N.~Goyal, M.~Artetxe, M.~Chen, S.~Chen, C.~Dewan, M.~Diab,
  X.~Li, X.~V. Lin, et~al.
\newblock {OPT}: Open pre-trained transformer language models.
\newblock {\em arXiv preprint arXiv:2205.01068}, 2022.

\bibitem{zhao2001using}
J.~Zhao.
\newblock Using dependence analysis to support software architecture
  understanding.
\newblock {\em arXiv preprint cs/0105009}, 2001.

\bibitem{zhao2023survey}
W.~X. Zhao, K.~Zhou, J.~Li, T.~Tang, X.~Wang, Y.~Hou, Y.~Min, B.~Zhang,
  J.~Zhang, Z.~Dong, et~al.
\newblock A survey of large language models.
\newblock {\em arXiv preprint arXiv:2303.18223}, 2023.

\end{thebibliography}

\end{document}